\documentclass[conference]{IEEEtran}
\usepackage{blindtext}
\usepackage{adjustbox}
\usepackage{multirow}
\usepackage{cite}
\usepackage{amsmath,amssymb,amsfonts}
\usepackage{algorithm,algpseudocode}
\algrenewcommand\algorithmicindent{0.5em}
\usepackage{wrapfig,lipsum,booktabs}
\usepackage{subfigure}
\usepackage{float}  
\usepackage{graphicx}
\usepackage{textcomp}
\usepackage{xcolor}
\usepackage{graphicx}
\usepackage{subcaption}
\usepackage{hyperref}
\usepackage[compact]{titlesec}

\let\svthefootnote\thefootnote
\newcommand\freefootnote[1]{%
  \let\thefootnote\relax%
  \footnotetext{#1}%
  \let\thefootnote\svthefootnote%
}

\DeclareRobustCommand{\IEEEauthorrefmark}[1]{\smash{\textsuperscript{\footnotesize #1}}}

\makeatletter
\algnewcommand{\LineComment}[1]{\Statex \hskip\ALG@thistlm \(\triangleright\) #1}
\makeatother

\usepackage[mathlines,switch]{lineno}

\makeatletter

\let\old@ps@IEEEtitlepagestyle\ps@IEEEtitlepagestyle
\def\confheader#1{%
    \def\ps@IEEEtitlepagestyle{%
        \old@ps@IEEEtitlepagestyle%
        \def\@oddhead{\strut\hfill#1\hfill\strut}%
        \def\@evenhead{\strut\hfill#1\hfill\strut}%
    }%
    \ps@headings%
}
\makeatother

\confheader{%
    \parbox{10cm}{\centering 2024 IEEE International Conference on Big Data (BigData)}}
    
\begin{document}

\title{SplitVAEs: Decentralized scenario generation from siloed data for stochastic optimization problems}

\author{
\IEEEauthorblockN{H M Mohaimanul Islam\IEEEauthorrefmark{1},
Huynh Q. N. Vo\IEEEauthorrefmark{1},
Paritosh Ramanan\IEEEauthorrefmark{1}}
\IEEEauthorblockA{\IEEEauthorrefmark{1}School of Industrial Engineering and Management}
\IEEEauthorblockA{Oklahoma State University, Stillwater, OK 74078 USA}
\IEEEauthorblockA{Email: \{h\_m\_mohaimanul.islam, lucius.vo, paritosh.ramanan\}@okstate.edu}
}

\maketitle

\begin{abstract}
Stochastic optimization problems in large-scale multi-stakeholder networked systems (e.g., power grids and supply chains) rely on data-driven scenarios to encapsulate complex spatiotemporal interdependencies. However, centralized aggregation of stakeholder data is challenging due to the existence of data silos resulting from  computational and logistical bottlenecks. In this paper, we present SplitVAE, a decentralized scenario generation framework that leverages variational autoencoders to generate high-quality scenarios without moving stakeholder data. With the help of experiments on distributed memory systems, we demonstrate the broad applicability of SplitVAEs in a variety of domain areas that are dominated by a large number of stakeholders. Furthermore, these experiments indicate that SplitVAEs can learn spatial and temporal interdependencies in large-scale networks to generate scenarios that match the joint historical distribution of stakeholder data in a decentralized manner. Lastly, the experiments demonstrate that SplitVAEs deliver robust performance compared to centralized, state-of-the-art benchmark methods while significantly reducing data transmission costs, leading to a scalable, privacy-enhancing alternative to scenario generation.
\end{abstract} 


\section{Introduction}\label{sec:intro}\freefootnote{This work was supported in part by the National Science Foundation's SaTC program under grant CNS-2348411. The computing for this project was performed at the High Performance Computing Center at Oklahoma State University supported in part through the National Science Foundation grant OAC-1531128. (Corresponding author: H M Mohaimanul Islam).}
Stochastic optimization (SO) problems are essential components in large-scale, data-driven planning and decision frameworks for multi-stakeholder-led networked systems including but not limited to energy \cite{zhao2015data}, supply chain \cite{chopra2007supply}, spare parts management \cite{shi2023stochastic}, healthcare \cite{anderson2014stochastic} and logistics \cite{erera2010vehicle}. Good quality solutions to SO problems fundamentally require high-fidelity data-driven scenarios that effectively encapsulate spatial and temporal interdependencies from myriad sources \cite{doi:10.1137/1.9781611973433}. Consequently, the generation of high-fidelity scenarios is heavily contingent on the availability of high-quality datasets that capture these interdependencies from heterogeneous data sources. Conventional scenario generation methods necessitate centralized data aggregation to capture interdependencies to drive stochastic planning decisions. 
These centralized aggregation schemes, however, face several computational, and logistical challenges \cite{li2021privacy,goncalves2021privacy} due to the presence of data silos across multiple stakeholders of a networked system. 
Therefore, in this study, we present SplitVAE, a novel decentralized scenario generation framework leverages split-learning to train variational autoencoders (VAEs) to eliminate the demand for centralized data aggregation. SplitVAEs are compatible with pre-existing data silos, ensure scalable generation of high-fidelity scenarios and significantly reduce data transfer costs.

\begin{figure}[htbp]
    \centering
    \includegraphics[width=\linewidth,keepaspectratio]{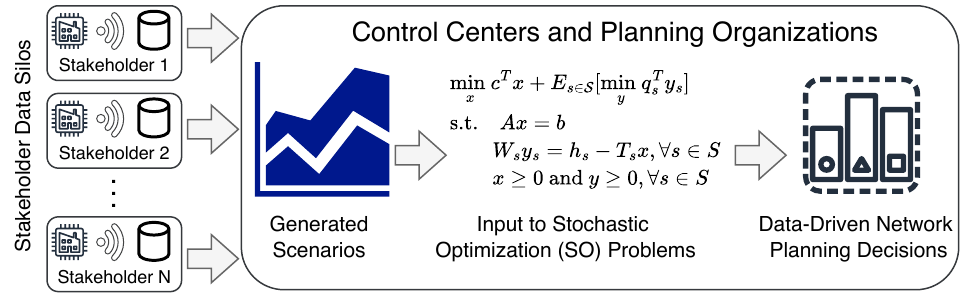}
    \caption{Scenario generation for stochastic optimization.}
    \label{fig:stochasticoptimization}
\end{figure}

Conventionally, SO solution frameworks rely on the acquisition of local spatiotemporally correlated datasets from individual stakeholders. The mechanism of data aggregation is typically carried out by a control center or planning organization as represented in Figure \ref{fig:stochasticoptimization}. Such coordinating entities are typically found in multi-stakeholder networked systems, for instance, Independent System Operators (ISOs) in power systems and control towers in the supply chain. The aggregated dataset is then used to generate high-fidelity scenarios that can be used to solve an SO problem in a centralized fashion at the control center level. The solutions to these SO problems represent data-driven planning decisions that can be in turn used to determine an operational schedule for each stakeholder subject to constraints like reliability, safety and demand satisfaction. However, the entire data driven planning mechanism is challenged in cases where data silos are present across various stakeholders.

Therefore, the goal of SplitVAEs is to enable planning organizations and control centers to generate generate stochastic decisions without incurring the burden of data processing and aggregation from multiple system stakeholders. Such planning organizations typically have regulatory oversight on the operational aspects of these systems necessitating SO problems to be solved in a high-frequency fashion. Thus, our proposed SplitVAE framework relies on distributed memory based aggregation primitives to conduct training without the need to move local data. As a result, SplitVAEs by their very design are fully capable of functioning in the presence of geographically distributed, decentralized data silos as is typically found in multi-stakeholder infrastructure systems. 

As a result, SplitVAEs can potentially be extended to operate in hybrid edge-cloud environments based on containerized applications enabling significant reductions in data transmission. Additionally, it is capable of succinctly capturing spatiotemporal interdependencies in a decentralized manner despite the inherent non-linearities in the underlying data among multiple stakeholder subsystems while maintaining high scalability. SplitVAEs complement existing stochastic optimization frameworks by supplying high-fidelity scenarios generated from siloed datasets. SplitVAEs are not meant to be a replacement of stochastic optimization solution methodologies. Instead, they help generate spatiotemporally interdependent scenarios from siloed or geographically dispersed datasets, which can then be used as input for solving stochastic optimization problems. As a result, we can concisely summarize the contributions of our study as follows:

\begin{itemize}
\item We design a computationally efficient learning paradigm
that combines edge-based autoencoders with a server driven variational autoencoder to jointly learn spatiotemporal interdependencies of siloed data.
\item We develop a scalable algorithmic framework that decomposes global backpropagation steps into edge and server-based sub-problems, enabling the bi-directional flow of learning insights, eliminating the movement of data.
\item We propose a distributed memory-based computational paradigm that can be seamlessly adopted to enable a scalable, real-world implementation of the SplitVAEs framework to generate scenarios for SO problems.
\item We demonstrate the ability of SplitVAEs to handle heterogeneous datasets across diverse areas by comparing the generated scenarios with established benchmark methods.
\end{itemize}

The remaining parts of this study are organized in the following manner. Section \ref{sec:related_works} provides a literature review of prior work. Sections \ref{sec:splitVAEs_design} and \ref{sec:splitVAEs_algorithm} describe the architecture and the algorithmic design of our proposed framework, respectively. Section \ref{sec:benchmark} provides a summary of benchmark methods as well as statistical evaluation metrics. Section \ref{sec:results} analyzes the performance of the SplitVAEs framework across different application domains and diverse computational settings. Section \ref{sec:conclusion} serves as the concluding section of this study.

\section{Related Works}\label{sec:related_works}
There have been several scenario generation mechanisms that have been proposed including auto-regressive integrated moving average (ARIMA) methods \cite{morales2010methodology,papavasiliou2011reserve}, Gaussian copula-based approaches \cite{pinson2009probabilistic,werho2021scenario} and deep generative adversarial networks (GANs) \cite{liang2019sequence,li2021privacy}. A majority of these studies, however, dealt with data centralization leading to computational and data privacy concerns \cite{li2021privacy}. From a computational standpoint, centralized methods such as ARIMA and copula-based approaches rely on the construction and factorization of covariance matrices \cite{papavasiliou2015stochastic, werho2021scenario} aided by the accumulation of stakeholder data. Meanwhile, centralized GAN-based methods suffer from high instability and challenges in convergence \cite{saxena2021generative} during model training. As a result, scalability remains an issue with these methods, especially for large-scale networks and systems consisting of myriad sources of uncertainty.

Further, centralized methods depend on nonlinear modeling of relationships between predictor feature variables and the generated scenarios that are connected in space and time. The authors in \cite{papavasiliou2015stochastic} used wind speed data to construct scenarios of wind power generation via the power curve of wind turbines. The work in \cite{rios2015multi} leveraged the nonlinear relationship between temperature and power consumption using epi-spline functions to generate scenarios. These approaches assume homogenous sources of data. When integrating heterogeneous sources of data such as wind, PV farms, and alternative fuel sources like hydrogen, these approaches might become infeasible.

On the other hand, VAEs \cite{Kingma_2019} serve as a potential alternative to methods that rely on non-linear modeling requirements. VAEs can help synthesize new, previously unseen scenarios by capturing the distribution of more concise, low-dimensional embeddings in a latent space. By sampling from the learned latent space distributions, VAEs can be used to generate scenarios that are spatially and temporally interdependent. Unlike GANs however, the VAEs provide a scalable, and decomposable implementation paradigm which is suited for data silos with heterogenous features as well.

Additionally, challenges with respect to siloed data arise in several application domains. For example, visibility across stakeholder subsystems is a major hindrance in supply chain and logistics owing to heterogeneity and incompatibility of local data storage systems \cite{somapa2018characterizing}. Similarly, in the medical field, data privacy is the primary reason for the existence of data silos \cite{antunes2022federated}. Privacy obstacles to scenario generation could potentially stem from the sensitivity of underlying datasets from the commercial standpoint \cite{li2023wind}. Such data sharing challenges are especially prevalent in the case of variable renewable energy (VRE) stakeholders \cite{jia2018distributed,goncalves2021privacy}. 
Overall, conventional methods of scenario generation face several limitations in terms of their applicability to heterogeneous data and requirements for privacy preservation from diverse stakeholders.  

  
Split learning \cite{vepakomma2018split,singh2019detailed,poirot2019split} presents a more generalized approach to handling siloed data sources in a decentralized, privacy-friendly fashion. Instead of relying on horizontal partitioning of data, split learning relies on model partitioning schemes that yield low-dimensional embeddings from heterogeneous data. Split learning paradigms leverage these embeddings to learn the interdependencies of the features contained in the siloed heterogeneous datasets without the need to move local data. As a result, split learning methods can potentially help resolve the data heterogeneity and data silo challenges that are inherent in scenario generation methods. Thus, our approach combines the split learning paradigm with the VAE to yield a computationally efficient, privacy-enhancing framework capable of delivering high-quality scenarios from decentralized and siloed datasets.

\begin{figure*}[htbp]
	\centering
\includegraphics[width=\textwidth,keepaspectratio]{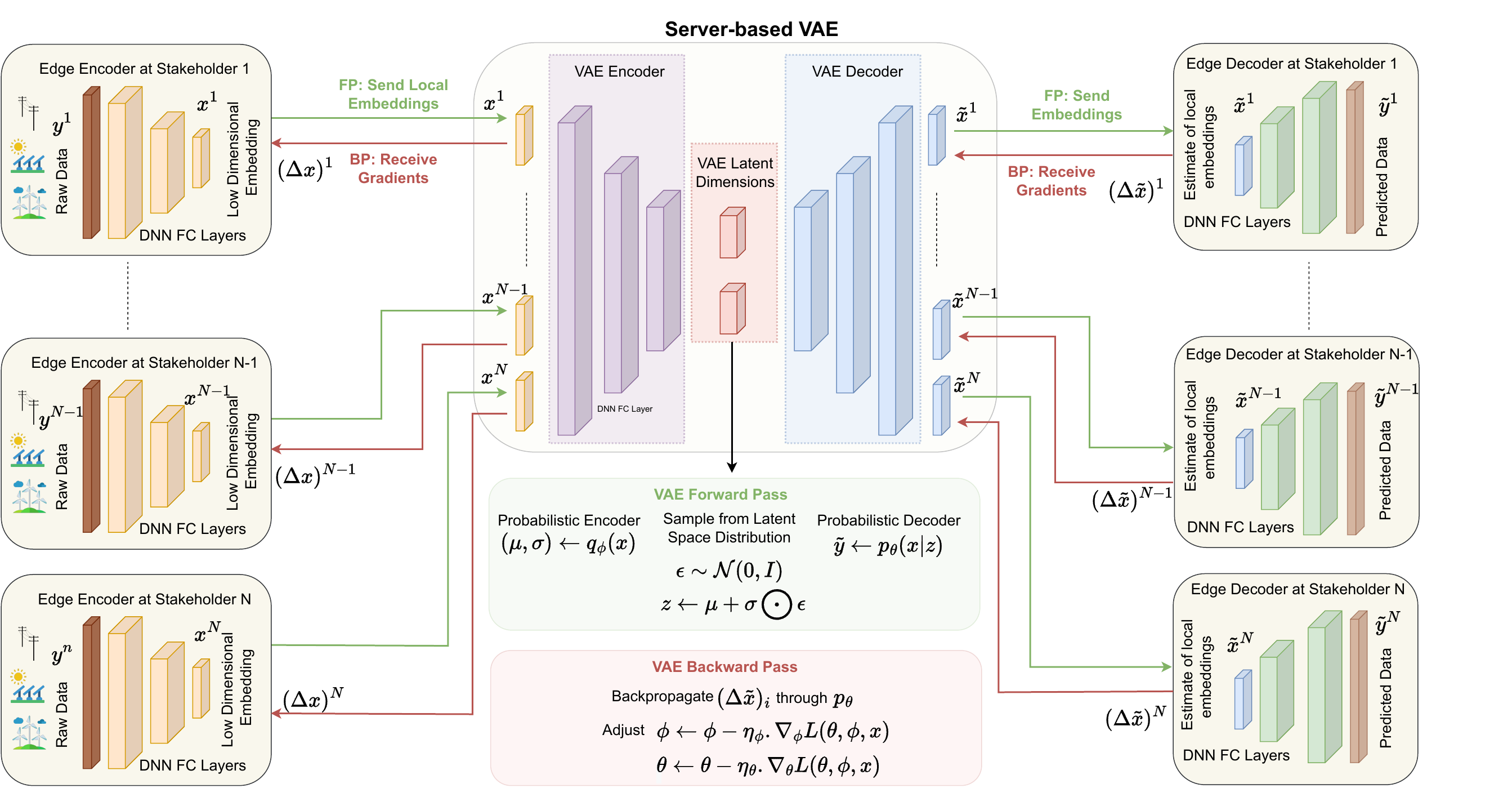}
	\caption{The learning paradigm of SplitVAEs.}\label{fig:SplitVAES}
\end{figure*}

\section{SplitVAEs Framework Design }\label{sec:splitVAEs_design}
We consider the SplitVAEs framework from the perspective of edge and server-level agents. The edge level represents heterogeneous system stakeholders that govern individual data silos comprising time series datasets. In our framework, the server entity houses the VAE and represents a planning organization, coordination agency, or a control center, while the edge entities consist of an autoencoder (AE) that is jointly trained in conjunction with the VAE at the server level. We begin our discussion by describing the conventional, centralized VAE (Central-VAE) formulation.

\subsection{Centralized variational autoencoder (VAE) formulation}\label{subsec:cvae}
The Central-VAE model consumes data provided by edge-level AEs, denoted by $x\in\mathbb{R}^d$, to yield latent space embeddings, denoted by $z\in\mathbb{R}^s$, with the help of a probabilistic encoder, denoted by $q^{\text{cvae}}_{\phi}(z|x)$. Specifically, $q^{\text{cvae}}_{\phi}(z|x)$ initially predicts the mean and variance of the latent space distribution, denoted by $(\mu_s,\sigma_s)$, as represented in Equation \eqref{eq:vae1}. The mean and variance pair $(\mu_s,\sigma_s)$ are then utilized to generate latent space representations $z$ with the help of sampling from the standard normal distribution as represented in Equation \eqref{eq:vae2}. The integration of a random sample, denoted by $\epsilon$, is referred to as the \textit{reparametrization trick} and is necessary to avoid computationally expensive simulation-driven latent space estimation \cite{Kingma_2019}. Similarly, a probabilistic decoder, denoted by $p^{\text{cvae}}_{\theta}(x|z)$, is applied so as to obtain a reconstruction, $\tilde{x}\in\mathbb{R}^d$, of the provided data $x$ represented by Equation \eqref{eq:vae3}.

\begin{gather}
(\mu_s,\sigma_s) \leftarrow q^{\text{cvae}}_{\phi}(x); \quad x \in \mathbb{R}^d \label{eq:vae1}\\
z \leftarrow \mu_s + \sigma_s \odot \epsilon, \text{ where } \epsilon \sim \mathcal{N}(0,I_s) \label{eq:vae2}\\
\tilde{x} \leftarrow p^{\text{cvae}}_{\theta}(z); \quad z\in\mathbb{R}^s, \tilde{x}\in\mathbb{R}^d \label{eq:vae3}
\end{gather}

Equations \eqref{eq:vae1} to \eqref{eq:vae3} are considered to be one forward pass of the VAE model. On the other hand, the backward pass of the VAE is carried out with the help of a total loss function represented by Equation \eqref{eq:vael1}. The total loss comprises of two errors: the reconstruction error, denoted by $L^{\text{recon}}(\theta,x,z)$ and indicated by Equation \eqref{eq:vaerecon}; the Kullback-Leibler divergence, denoted by $L^{\text{KL}}(\phi,\theta,x,z)$ and derived in \eqref{eq:vaeKL}, which measures the similarity of latent space distributions based on the probabilistic encoders and decoders.

\begin{gather}
L(\theta,\phi,x,z) = L^{\text{recon}}(\theta,x,z) + L^{\text{\text{KL}}}(\phi,\theta,x,z) \text{\quad} \label{eq:vael1}\\
L^{\text{recon}}(\theta,x,z) = -\mathbb{E}_{z \sim q^{\text{cvae}}_{\phi}(z | x)}\Big(\log p^{\text{cvae}}_{\theta}(x | z)\Big) \text{   } \label{eq:vaerecon}\\
L^{\text{\text{KL}}}(\phi,\theta,x,z) = D_{\text{KL}}\Big(q^{\text{cvae}}_{\phi}(z | x)\big| \big| p^{\text{cvae}}_{\theta}(z)\Big) \text{\quad\quad} \label{eq:vaeKL}
\end{gather}

Therefore, the implementation of the Central-VAE requires the transfer of edge-level stakeholder datasets to compute the reconstruction loss as well as generate latent space embeddings, which might be infeasible due to the reasons outlined in Section \ref{sec:intro}. Therefore, to learn the spatiotemporal interdependencies given the decentralized, siloed nature of stakeholder data, the SplitVAEs framework incorporates a novel split training mechanism that leverages an edge-based AE component and a server-based VAE component.

\subsection{Edge-level autoencoder (AE) formulation}
Given the aforementioned reasons, at the edge-level, we implement an AE-based framework comprising deep neural network (DNN) encoder and decoder models. We denote the siloed time series data located at each edge location $n \in \mathcal{N}$ as $y^n$. The functional forms of the encoder and decoder components are denoted by $x^n \leftarrow F_{\theta^n_E}(y^n)$ and $\tilde{y}^n \leftarrow G_{\theta^n_D}(x^n)$, respectively. For the edge-based AE framework, let $x^n,\tilde{y}^n,\theta^n_E,\theta^n_D$ be the low-dimensional embedding, the reconstructed vector, and the model parameters of the encoder and decoder, respectively. Unlike conventional AE training, a unique aspect of the SplitVAEs training mechanism is the independent training of the encoder and decoder components to holistically capture spatial and temporal interdependencies among all the edge devices. 

\subsection{Server-level VAE formulation}
At the server level, we implement a conventional VAE framework. However, unlike the Central-VAE, in each training epoch, the server model consumes the set of low dimensional embeddings, denoted by $\{x^1,\ldots, x^{N-1}, x^{N}\}$, supplied by the edge-based AEs at the various stakeholders. Let $q_{\phi}(z|x)$ be the probabilistic encoder, responsible for deriving the latent vector $z$ from the input embeddings $\{x\}$ in terms of empirical mean and standard deviation - denoted by $\hat{\mu}$ and $\hat{\sigma}$, respectively. Additionally, we denote $z\leftarrow R_{\hat{\mu},\hat{\sigma}}(\epsilon)$ as the reparametrization function akin to Equation \eqref{eq:vae2}. The second core component is the latent vector $z$ representing the compressed data in a lower dimension. The size of $z$ becomes a crucial hyperparameter that necessitates careful tuning, in conjunction with the overall architecture of the entire network. The final component is the probabilistic decoder, denoted by $p_{\theta}(x|z)$, responsible for reconstructing a data point $x$ from a given latent vector $z$. 

\subsection{Loss Functions}
Here, we discuss the relevant formulations of the reconstruction error and the KL loss that are utilized by the SplitVAEs framework for training.

\noindent
\textbf{Reconstruction Loss}: We employ the binary cross-entropy (BC) loss function at the $n$-{th} edge location for all $n \in \mathcal{N}$ to measure the mean reconstruction error between predicted values $\tilde{y}^n$ and observed values $y^n$ as represented in Equation \eqref{eq:bce_local}.

\begin{equation}\label{eq:bce_local}
L^{\text{BC}}_n = -\frac{1}{B}\sum\limits_{b=1}^{B}\Big[\tilde{y}^blog(y^b) + (1-\tilde{y}^b)log(1-y^b)\Big]
\end{equation}

In Equation \eqref{eq:bce_local}, let $B,\tilde{y}_b, y_b$ be the batch size, predicted, and target data for scenario generation, respectively. To preserve the simplicity of notations, we assume without loss of generality that the loss function $L^{\text{BC}}_n$ is computed with respect to flattened representations of local tensor data and estimates both $y^n$ and $\tilde{y}^n$ of dimension $d^n$. We note that the computation of $L^{\text{BC}}_n$ can be carried out independently at each edge location since it relies purely on siloed data, represented by $y^n$. 

\noindent
\textbf{KL Loss}: We utilize the formulation given in Equation \eqref{eq:kl_loss} to derive the relevant KL loss function as follows:

\begin{equation}\label{eq:kl_loss}
    L^{\text{KL}} = -\frac{1}{2}\Big[ 1 + 2 \log(\hat{\sigma}) - \hat{\mu}^2 - \hat{\sigma} \Big]
\end{equation}

In Equation \eqref{eq:kl_loss}, the computed KL loss depends only on the latent space embeddings resulting from the reparametrization trick described in Section \ref{subsec:cvae}. Consequently, the backpropagation of the error resulting from the KL loss can be initiated from the server level itself. Therefore, the decomposed reconstruction and KL error terms can be leveraged to effect a computationally efficient training mechanism of the SplitVAEs framework.

\section{SplitVAEs Algorithmic Framework}\label{sec:splitVAEs_algorithm}
There are two crucial characteristics of the SplitVAE decomposition scheme that can be advantageous to its training. First, the KL loss, $L^{\text{KL}}$, is purely a factor of the latent space embeddings estimated by the server-based VAE framework. Second, the reconstruction error can be computed in a fully decentralized fashion by each of the edge locations locally without revealing the siloed datasets themselves. Thus, only the local reconstruction loss needs to be back-propagated from the edge-based AE decoder to the server-based VAE decoder. This step can be immediately followed by a backpropagation of the total loss, $L^{\text{recon}} + L^{\text{KL}}$, from the server-level VAE encoder to the edge-based AE encoders. As a result of our two-pronged backpropagation mechanism, we can enable the flow of insights back and forth between edge and server-level devices, leading to seamless learning of the overall spatiotemporal interdependencies among stakeholders.

A single training epoch in our proposed framework can be divided based on edge and server level algorithmic frameworks. In the following sections, we discuss the decomposition of the forward and backward passes of the server-level VAE as well as the encoders and decoders of edge-level AEs. Specifically, without loss of generality, we consider a set of $\mathcal{N}$ edge nodes wherein $|\mathcal{N}|=N$. To accomplish the training of SplitVAEs, we leverage several common distributed memory aggregation primitives such as the scattering and gathering operations, denoted by \texttt{DM.Scatter} and \texttt{DM.Gather}, respectively. Let \texttt{Tensor.Concat, Tensor.Split}, and \texttt{BACKPROP} functions indicate tensor concatenation, split  and backpropagation primitives, respectively. We designate the server as the root process $r$, while the edge-level processes are denoted by $n$ and are similar to the concept of ranks in distributed memory frameworks. We use $k$ to denote iterative updates of the model parameters represented by $\theta^n_E, \theta^n_D,\phi,\theta$.

\subsection{Edge-based Algorithmic Framework}\label{subsubsec:edge}
One training epoch of the edge-based AE framework consists of the following computational steps.

\noindent
\textbf{AE Encoder Forward Pass}: The computational steps associated with the forward pass of the AE encoder are represented by the function \texttt{EdgeEncFP}. In \texttt{EdgeEncFP}, we consider local siloed training data $y^b_e$ that is passed through the AE encoder $F_{(\theta^n_E)^k}$ to yield the low dimensional embedding $(x^b_e)^n$. We utilize a \texttt{DM.Gather} operation to aggregate the embeddings at the server level. 

\begin{algorithm}[htbp]
\caption{AE Encoder Forward Pass}\label{alg:eefp}
    \begin{algorithmic}
        \Function{EncEdgeFP}{\textbf{root}:r, \textbf{agent}:n, \textbf{epoch}:$e$, \textbf{batch}:$b$}
        \State Acquire local data $y^b_e$
        \State Obtain $(x^b_e)^n \leftarrow F_{(\theta^n_E)^k}(y^b_e)$
        \State Execute \texttt{DM.Gather}\big(\textbf{root}:r, \textbf{data}:$(x^b_e)^n$\big) 
        \EndFunction
    \end{algorithmic}
\end{algorithm}

\noindent
\textbf{AE Decoder Forward Pass}: The function \texttt{EdgeDecFP} represents the reconstruction of siloed edge data. We leverage a \texttt{DM.Scatter} operation to collect the predicted values of the low-dimensional embedding outcomes of the server-level training denoted by $(\tilde{x}^b_e)^n$. These predictions are passed through the AE decoder $G_{\theta^{k}_D}$ to generate reconstructions and the BCE loss through Equation \eqref{eq:bce_local}.

\begin{algorithm}[htbp]
 \caption{AE Decoder Forward Pass}\label{alg:edfp}   
    \begin{algorithmic}
        \Function{EdgeDecFP}{\textbf{root}:r, \textbf{agent}:n, \textbf{epoch}:$e$, \textbf{batch}:$b$}
        \State $(\tilde{x}^b_e)^n\leftarrow$\texttt{DM.Scatter}(\textbf{root}:r, \textbf{data}:\texttt{NULL})
        \State Obtain $\tilde{y}^b_e \leftarrow G_{\theta^{k}_D}(\tilde{x}^b_e)$
        \State Compute reconstruction loss $({L_{BCE}})^b_e$ using \eqref{eq:bce_local}
        \EndFunction
    \end{algorithmic}

\end{algorithm}

\noindent
\textbf{AE Decoder Backward Pass}: As an immediate next step, our implementation leverages \texttt{EdgeDecBP} to back-propagate the errors based on the reconstruction losses. In \texttt{EdgeDecBP} $(g_{\theta_D})^b_e = \nabla_{\theta_D}(L^{\text{recon}}_n)^b_e$ represents the gradient of $(L^{\text{recon}}_n)^b_e$ concerning $\theta_D$ and $\Delta (\tilde{x})^b_e$ represents the error at the input layer of the edge-based decoder. Finally, we use a \texttt{DM.Gather} operation to collect the errors of the low-dimensional embeddings $(\tilde{x}^b_e)^n$ at the server.

\begin{algorithm}
\caption{AE Decoder Backward Pass}\label{alg:edbp}      
    \begin{algorithmic}
        \Function{EdgeDecBP}{\textbf{root}:r, \textbf{agent}:n, \textbf{epoch}:$e$, \textbf{batch}:$b$}
        \State Obtain $(\Delta \tilde{x})^b_e$
        \State Obtain $(g_D)^b_e\leftarrow$ \texttt{BACKPROP}\big($G_{(\theta^n_D)^k},(L^{^\text{BC}}_n)^b_e$\big)
        \State Update $\theta^{k+1}_D \leftarrow \theta^k_D - \eta_{D}.(g_D)^b_e$
        \State \texttt{DM.Gather}\big(\textbf{root}:r, \textbf{data}:$(\Delta \tilde{x}^b_e)^n$\big) 
        \EndFunction
    \end{algorithmic}
\end{algorithm}

\noindent
\textbf{AE Encoder Backward Pass}: The last step of the edge based training epoch is presented in \texttt{EdgeEncBP}. We leverage \texttt{DM.Scatter} operation that collects errors of low dimensional embeddings $(\Delta{x}^b_e)^n$ computed by the server. Consequently, the AE encoder back-propagates these errors through the local encoder $F_{(\theta^n_E)^k}$ to complete the training epoch.

\begin{algorithm}
\caption{AE Encoder Backward Pass}\label{alg:eebp}          
    \begin{algorithmic}
        \Function{EdgeEncBP}{\textbf{root}:r, \textbf{agent}:n, \textbf{epoch}:$e$, \textbf{batch}:$b$}
        \State $(\Delta{x}^b_e)^n\leftarrow$\texttt{DM.Scatter}(\textbf{root}:r, \textbf{data}:\texttt{NULL})
        \State $(\Delta y)^b_e,(g_{E})^b_e\leftarrow$\texttt{BACKPROP}\big($F_{(\theta^n_E)^k},\Delta{x}^b_e$\big)
        \State Update $(\theta^n_E)^{k+1} \leftarrow (\theta^n_E)^k - \eta_{E}.(g_E)^b_e$
        \EndFunction
    \end{algorithmic}
\end{algorithm}

\subsection{Server-based Algorithmic Framework}\label{subsubsec:server}
At the server, a training epoch corresponding to the VAE can be described based on the following components. 

\noindent
\textbf{VAE Forward Pass}: The forward pass of the server VAE framework is represented by \texttt{VAEServerFP}. Specifically, the server concatenates the low-dimensional embeddings obtained as a consequence of the \texttt{DM.Scatter} operation. The resulting concatenated embedding $x^b_e$ is passed through the probabilistic encoder $q_{\phi}$ and is reparametrized with the standard normal distribution to yield the latent space representation $z$. Subsequently, we pass $z$ through the probabilistic decoder to yield estimated low-dimensional embeddings, $\tilde{x}^b_e$, which are split into $N$ components according to the mandated dimensions of each edge agent.

\noindent
\textbf{VAE Backward Pass}: The backward pass of the VAE is represented by \texttt{VAEServerBP}. The backpropagated set of error values, denoted by $\Big[(\Delta \tilde{x}^b_e)^1, (\Delta \tilde{x}^b_e)^2, \ldots, (\Delta \tilde{x}^b_e)^N\Big]$ resulting from $L^{\text{recon}}_n$ are acquired from each edge agent through \texttt{DM.Gather}. These error terms are concatenated and back-propagated through the probabilistic decoders to yield $(\Delta z)^b_e$ and $(g_{\theta})^b_e$ representing latent space error terms and probabilistic decoder gradients, respectively. With a subsequent backpropagation of $(\Delta z)^b_e$ through the probabilistic encoder $q_{\phi}$, we obtain $(\Delta x^{^\text{BC}})^b_e$ and $(g^{^\text{BC}}_{\phi})^b_e$ corresponding to the error terms at the input leaves of the VAE and the gradient values for the probabilistic encoder, respectively. More importantly, the errors and gradients accumulated so far on the VAE correspond to the reconstruction error terms observed at the edge level only. 

\begin{algorithm}[!htbp]
\caption{VAE Forward Pass}\label{alg:vsfp}
    \begin{algorithmic}
        \Function{VAEServerFP}{\textbf{root}:r, \textbf{epoch}:$e$, \textbf{batch}:$b$}
        \State $\Big[(x^b_e)^1, \ldots, (x^b_e)^N\Big]\leftarrow$
        \texttt{DM.Gather}(\textbf{root}:r,\textbf{data}:\texttt{NULL})
        \State Acquire $x^b_e\leftarrow$ \texttt{Tensor.Concat}$\big[(x^b_e)^1, (x^b_e)^2, \ldots, (x^b_e)^N\big]$
        \State Obtain $(\hat{\mu}^b_e,\hat{\sigma}^b_e) \leftarrow q_{\phi}(x^b_e)$
        \State Create $z^b_e \leftarrow \hat{\mu}^b_e + \hat{\sigma}^b_e \odot \epsilon^b_e$, where $\epsilon^b_e \sim \mathcal{N}(0,I_s)$
        \State Estimate $\tilde{x}^b_e \leftarrow p_{\theta}(z^b_e)$
        \State $\Big[(\tilde{x}^b_e)^1, (\tilde{x}^b_e)^2 \ldots (\tilde{x}^b_e)^N\Big]\leftarrow$\texttt{Tensor.Split}$\big[\tilde{x}^b_e\big]$
        \State \texttt{DM.Scatter}(\textbf{root}:r, \textbf{data}:$\Big[(\tilde{x}^b_e)^1, \ldots, (\tilde{x}^b_e)^N\Big]$)
        \State Compute KL loss $(L^{\text{KL}})^b_e$ using \eqref{eq:kl_loss}
        \EndFunction
    \end{algorithmic}
\end{algorithm}

However, to complete VAE training, we would also need to incorporate the KL loss. Since the KL loss is purely a function of the latent space embeddings $z^b_e$, we backpropagate $(L^{\text{KL}})^b_e$ computed during \texttt{VAEServerFP} and accumulate additional error values at the input leaves. The final set of accumulated errors $(\Delta x)^b_e$ are split and sent back to the corresponding edge agents using \texttt{DM.Scatter}.

\subsection{Decentralized Training Algorithm}
Combining the training steps of the edge and server-level models provides us with Algorithm \ref{alg:vae} and \ref{alg:ae} for VAE and AE, respectively. In these algorithms, we abstract the training mechanism using the set of functions detailed in Sections \ref{subsubsec:edge} and \ref{subsubsec:server} for a specific set of batch and epoch sizes.

\begin{algorithm}[htbp]
\caption{VAE Backward Pass}\label{alg:vsbp}    
    \begin{algorithmic}
    \Function{VAEServerBP}{\textbf{root}:r, \textbf{epoch}:$e$, \textbf{batch}:$b$}
    \State $\Big[(\Delta \tilde{x}^b_e)^1,\ldots,(\tilde{x}^b_e)^N\Big]\leftarrow$\texttt{DM.Gather}(\textbf{root}:r,  \textbf{data}:\texttt{NULL})
    \State $\Delta{\tilde{x}}^b_e\leftarrow$ \texttt{Tensor.Concat}$\big[(\Delta \tilde{x}^b_e)^1, \ldots, (\Delta \tilde{x}^b_e)^N\big]$
    
    \State \LineComment{Backpropagate BC loss:}
    \State $(\Delta z)^b_e,(g_{\theta})^b_e\leftarrow$\texttt{BACKPROP}\big($p_{\theta},\Delta{\tilde{x}}^b_e$\big)
    \State $\Big[(\Delta \hat{\mu})^b_e, (\Delta \hat{\sigma})^b_e\Big]^{\text{BC}} \leftarrow$\texttt{BACKPROP}\big($R_{\hat{\mu},\hat{\sigma}},(\Delta z)^b_e$\big)
    \State $(\Delta x^{^\text{BC}})^b_e,(g^{^\text{BC}}_{\phi})^b_e\leftarrow$\texttt{BACKPROP}\big($q_{\phi},\Big[(\Delta \hat{\mu})^b_e, (\Delta \hat{\sigma})^b_e\Big]^{\text{BC}}$\big)
    \State \LineComment{Backpropagate KL loss:}
    \State $[(\Delta \hat{\mu})^b_e, (\Delta \hat{\sigma})^b_e]^{\text{KL}} \leftarrow$
    \texttt{BACKPROP}\big($R_{\hat{\mu},\hat{\sigma}},{L_{\text{KL}}}^b_e$\big)
    \State $(\Delta x^{\text{KL}})^b_e,(g^{\text{KL}}_{\phi})^b_e\leftarrow$\texttt{BACKPROP}\big($q_{\phi},\Big[(\Delta \hat{\mu})^b_e, (\Delta \hat{\sigma})^b_e\Big]^{\text{KL}}$\big)
    
    \State \LineComment{Update parameters}
    \State $\theta^{k+1} \leftarrow \theta^k -\eta_{\theta}.(g_{\theta})^b_e$
    \State $\phi^{k+1} \leftarrow \phi^k -\eta_{\phi}.\nabla_{\phi}((g^{^\text{BC}}_{\phi})^b_e + (g^{\text{KL}}_{\phi})^b_e)$
    \State \LineComment{Accumulate errors at leaves}
    \State $(\Delta x)^b_e \leftarrow(\Delta x^{\text{KL}})^b_e + (\Delta x^{^\text{BC}})^b_e$
    \State \LineComment{Scatter errors}
    \State $\Big[(\Delta{x}^b_e)^1,\ldots, (\Delta{x}^b_e)^N\Big]\leftarrow$
    \texttt{Tensor.Split}$\big[ \Delta{x}^b_e\big]$
    \State \texttt{DM.Scatter}(\textbf{root}:r, \textbf{data}:$\Big[(\Delta{x}^b_e)^1, (\Delta{x}^b_e)^2, \ldots, (\Delta{x}^b_e)^N\Big]$
    \EndFunction
    \end{algorithmic}
\end{algorithm}

The sequence of computation and communication steps outlined in Algorithms \ref{alg:vae} and \ref{alg:ae} is represented in Figure \ref{fig:SplitVAES}. The algorithmic architecture mentioned in this section enables planning organizations, regulators, and control centers to generate high-quality scenarios by sampling from the latent distribution space. 

\begin{algorithm}[!htbp]
\caption{Training Mechanism of Server-based VAE}\label{alg:vae}
\begin{algorithmic}
\For{b=0,1,2,\ldots \textbf{BatchSize}}
\For{i=0,1,2,\ldots \textbf{MaxEpochs}}
\State \Call{VAEServerFP}{root:r,epoch:e,batch:b}
\State \Call{VAEServerBP}{root:r,epoch:e,batch:b}
\EndFor
\EndFor
\end{algorithmic}
\end{algorithm}

\begin{algorithm}
\caption{Training Mechanism of Edge-based Autoencoder}\label{alg:ae}
    \begin{algorithmic}
        \For{b=0,1,2,\ldots \textbf{BatchSize}}
        \For{i=0,1,2,\ldots \textbf{MaxEpochs}}
        \State \Call{EdgeEncFP}{root:r,agent:n,epoch:e,batch:b}
        \State \Call{EdgeDecFP}{root:r,agent:n,epoch:e,batch:b}
        \State \Call{EdgeDecBP}{root:r,agent:n,epoch:e,batch:b}
        \State \Call{EdgeEncBP}{root:r,agent:n,epoch:e,batch:b}
        \EndFor
        \EndFor
    \end{algorithmic}
\end{algorithm}

From an implementation and deployment perspective, we envision a scenario where the server can generate a latent space sample to generate the low-dimensional embeddings for different edge locations by using a pre-trained VAE. The authority at the server level can now request the scenarios from each edge location corresponding to the provided latent space estimate. As a result, the server-based entity can effectively gather a sufficient number of scenarios so as to facilitate the solution of the SO problem.

\section{Benchmark Methods}\label{sec:benchmark}
In this section, we discuss the relevant benchmarking strategies for SplitVAEs by outlining the evaluation methods as well as the evaluation metrics employed in this
study.

\subsection{Evaluation Methods}
We compare SplitVAEs with centralized scenario generation methods, including the Gaussian copula \cite{electronics12071601,houssou2022generation} and the Centralized-VAE. The Gaussian copula models the joint probability distribution function (PDF) to capture the interdependent structures, specifically the spatiotemporal interdependencies within the data. Meanwhile, in SplitVAEs and Centralized-VAE, these interdependencies are modeled within the latent space $z$ when training the probabilistic encoder $q_{\phi}(z|x)$ and decoder $p_{\theta}(x|z)$ pair, as mentioned in Section \ref{sec:splitVAEs_design}. Further, given that the datasets and their corresponding scenarios are multidimensional in nature, we employ t-SNE graphs\cite{Hinton_2008} to visualize kernel density distributions of the observed and generated data, which are transformed into one-dimensional embeddings.

\subsection{Evaluation Metrics}
In addition to visual analyses, we employ multiple quantitative evaluation metrics to evaluate the quality of the generated scenarios. For each metric, lower values indicate a better model performance.

Let $X\in\mathbb{R}^{m_1\times d}$ be the real data and $Y\in\mathbb{R}^{m_2\times d}$ be the generated scenarios, each comprising of $d$ features with $m_1, m_2$ number of observations respectively. Thus, $X$ and $Y$ can be expressed as: $X=\{x_1,x_2,...,x_{m_1}\}$, where $x_i\in\mathbb{R}^d$ for all $ 1 \leq i \leq m_1$; $Y=\{y_1,y_2,...,y_{m_2}\}$, where $y_i\in\mathbb{R}^d$ for all $1 \leq i \leq m_2$. Hence, we can derive each metric as follows:

\vspace{1mm}
\noindent
\textbf{Fréchet inception distance (FID)}: The FID score is a widely accepted, state-of-the-art metric employed to evaluate deep generative models (DGMs) \cite{DBLP:journals/corr/abs-1812-04948, DBLP:journals/corr/abs-1912-04958}. The FID measures the Wasserstein distance between two PDFs $f_X(x)$ and $f_Y(y)$. In our study, the observed dataset and the generated scenarios, have finite values pertaining to their respective mean and covariance matrices denoted by $\mu_x,\mu_y$ and $\Sigma_x,\Sigma_y$ respectively. Under such conditions, based on \cite{DOWSON1982450} the FID can be defined as follows:

\vspace{-3mm}
\begin{equation}
    d_F[X,Y] = ||\mu_x - \mu_y||_2^2 - \text{Tr}\Big(\Sigma_x + \Sigma_y - 2\big(\Sigma_x\cdot\Sigma_y\big)^{\frac{1}{2}}\Big)
\end{equation}


\noindent
\textbf{Energy Score}: The energy score (ES) measures the dissimilarity in probability density between two multivariate random variables  \cite{SZEKELY20131249}, which are the observed data and generated scenarios in our study:

\vspace{-3mm}
\begin{equation}
    \epsilon_{m_1,m_2}(X,Y) = \frac{1}{m_1m_2}\sum_{i=1}^{m_1}\sum_{j=1}^{m_2}|x_i - y_j| - \frac{1}{2n_1^2}\sum_{i=1}^{m_1}\sum_{j=1}^{m_1}|x_i - x_j|
\end{equation}

\noindent
\textbf{Root-mean-square Error}: The well-known root-mean-square error (RMSE) measures the quadratic mean of the differences between the observed values and generated ones:

\vspace{-1mm}
\begin{equation}
    \text{RMSE} = \sqrt{\frac{1}{kd}\sum_{i=1}^{k}\sum_{j=1}^d\Big(x_{i,j} - y_{i,j}\Big)^2}
\end{equation}

\vspace{-1.5mm}
where $k = \min(m_1,m_2)$.

\textbf{Continuous ranked probability score}: The CRPS \cite{doi:10.1287/mnsc.22.10.1087} measures the squared distance between a single multivariate observed observation $y$ and the median of $F(x)$ which is the empirical cumulative distribution of generated scenarios \cite{bjerregard_introduction_2021}. For $m_2$ number of observations, the CRPS is given as:


\vspace{-3mm}
\begin{equation}
    \text{CRPS}(F,\{y_1,...,y_{m_2}\}) = \frac{1}{m_2}\sum_{i=1}^{m_2}\int\limits^{\infty}_{-\infty}\Big(F(x) - \mathbf{1}(x\geq y_i)\Big)^2dx
\end{equation}

\begin{figure*}[!htbp]
    \centering
        \subfigure[USAID]{\includegraphics[width=0.24\textwidth,keepaspectratio]{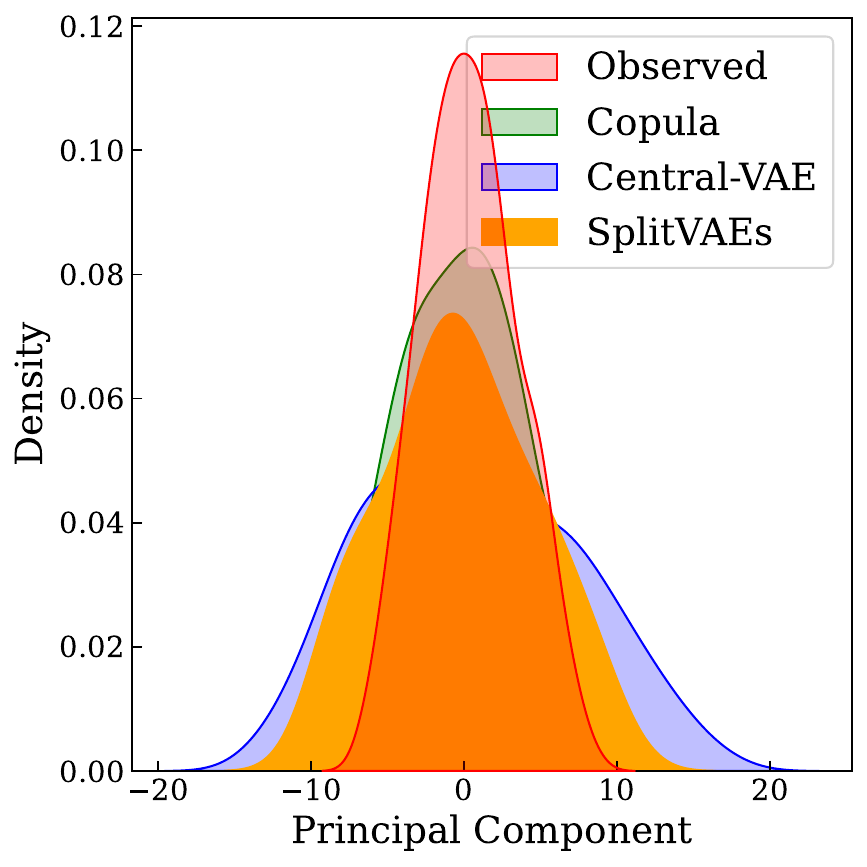}} 
        \subfigure[ACES]{\includegraphics[width=0.24\textwidth,keepaspectratio]{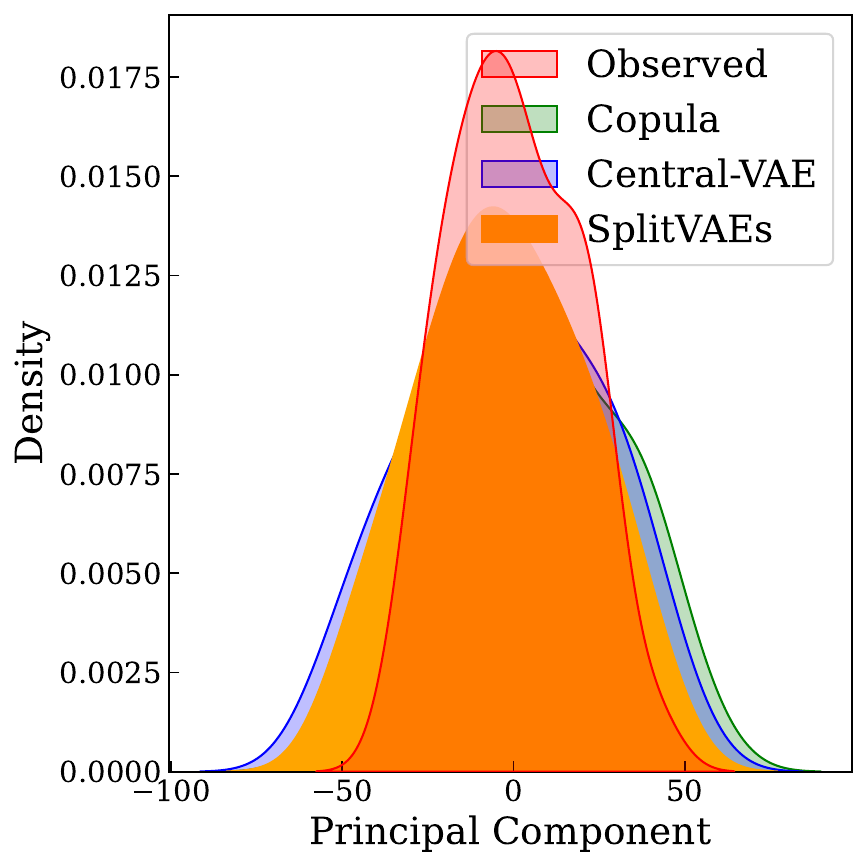}}
        \subfigure[Demand]{\includegraphics[width=0.24\textwidth,keepaspectratio]{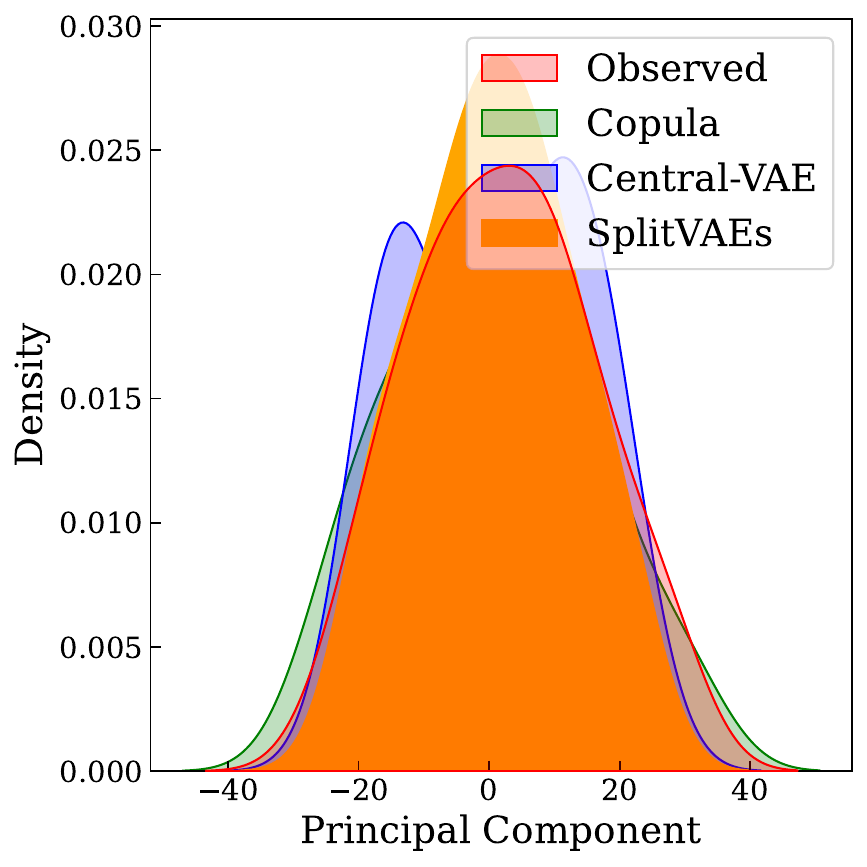}}
        \subfigure[Renewable]{\includegraphics[width=0.24\textwidth,keepaspectratio]{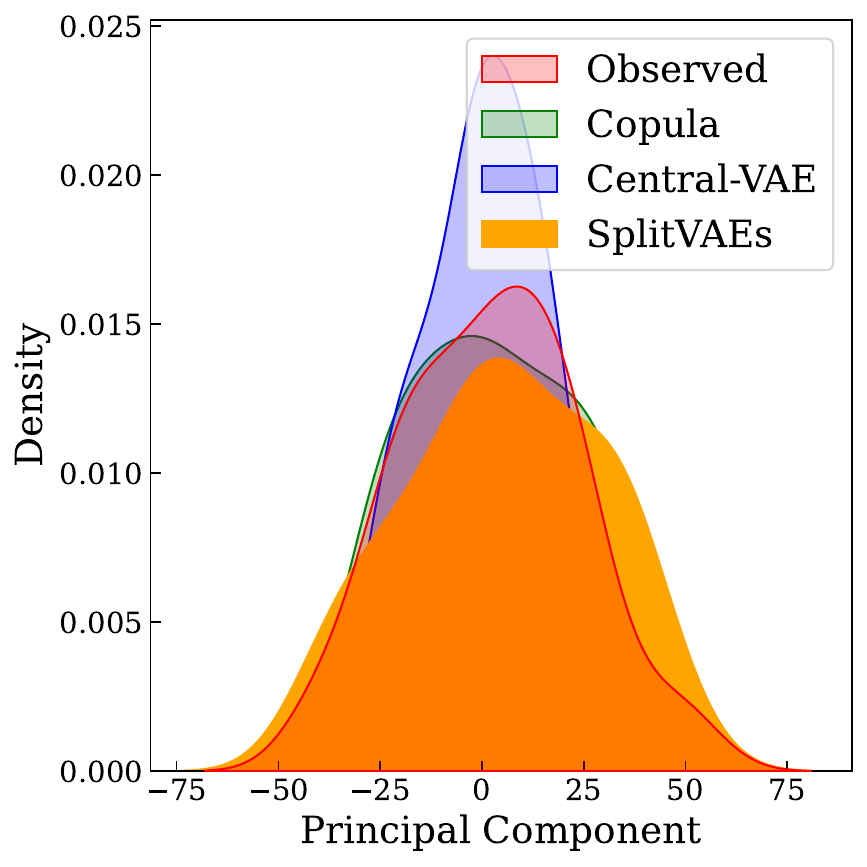}}
  \caption{Embedding distributions between observed data and generated scenarios using different methods across four cases.}\label{fig:t-SNE}
  \vspace{-3mm}
\end{figure*}
\begin{figure*}[htbp]
    \centering
        \subfigure[ACES time series]{\includegraphics[width=0.32\textwidth,keepaspectratio]{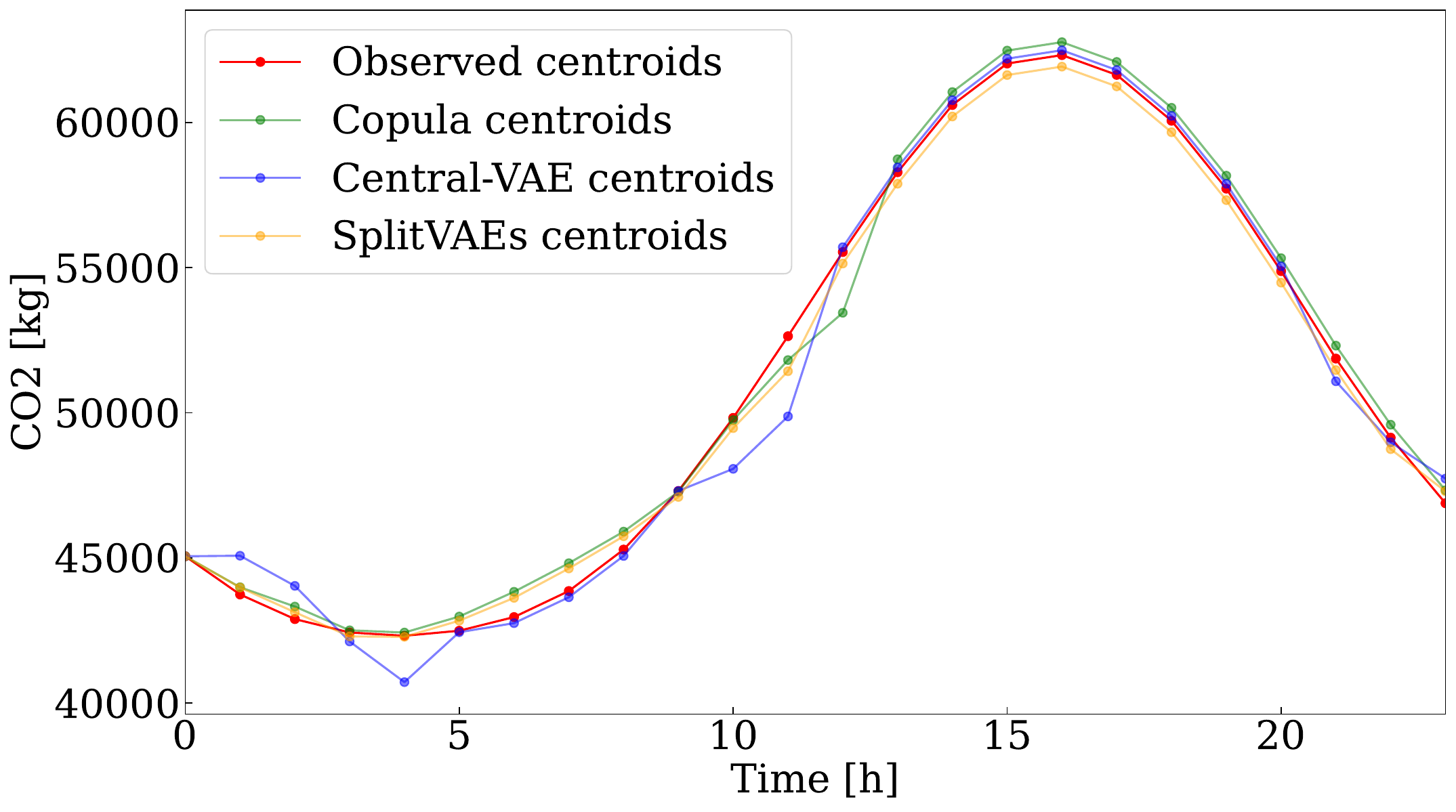}}
        \subfigure[Demand time series]{\includegraphics[width=0.32\textwidth,keepaspectratio]{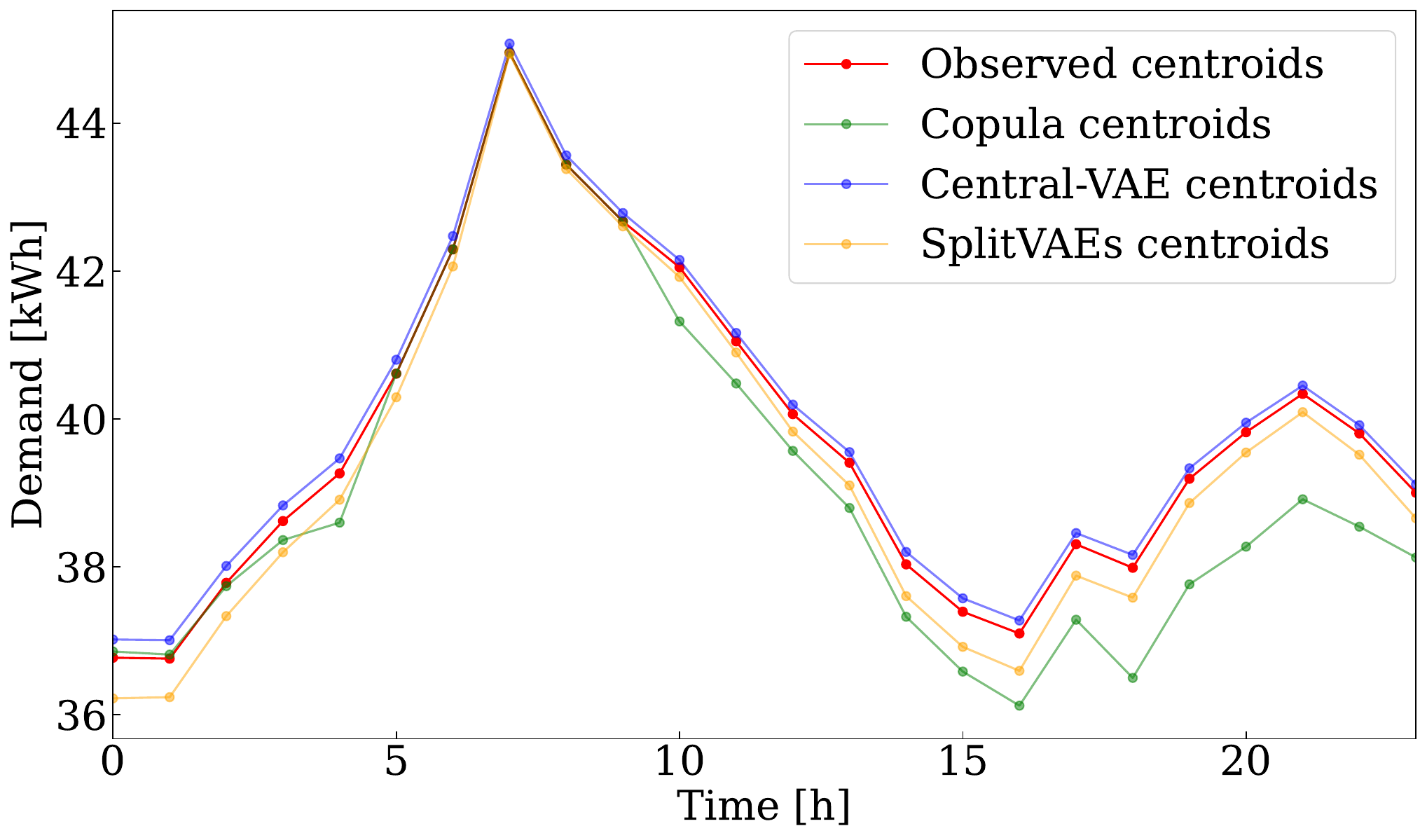}}
        \subfigure[Renewable time series]{\includegraphics[width=0.32\textwidth,keepaspectratio]{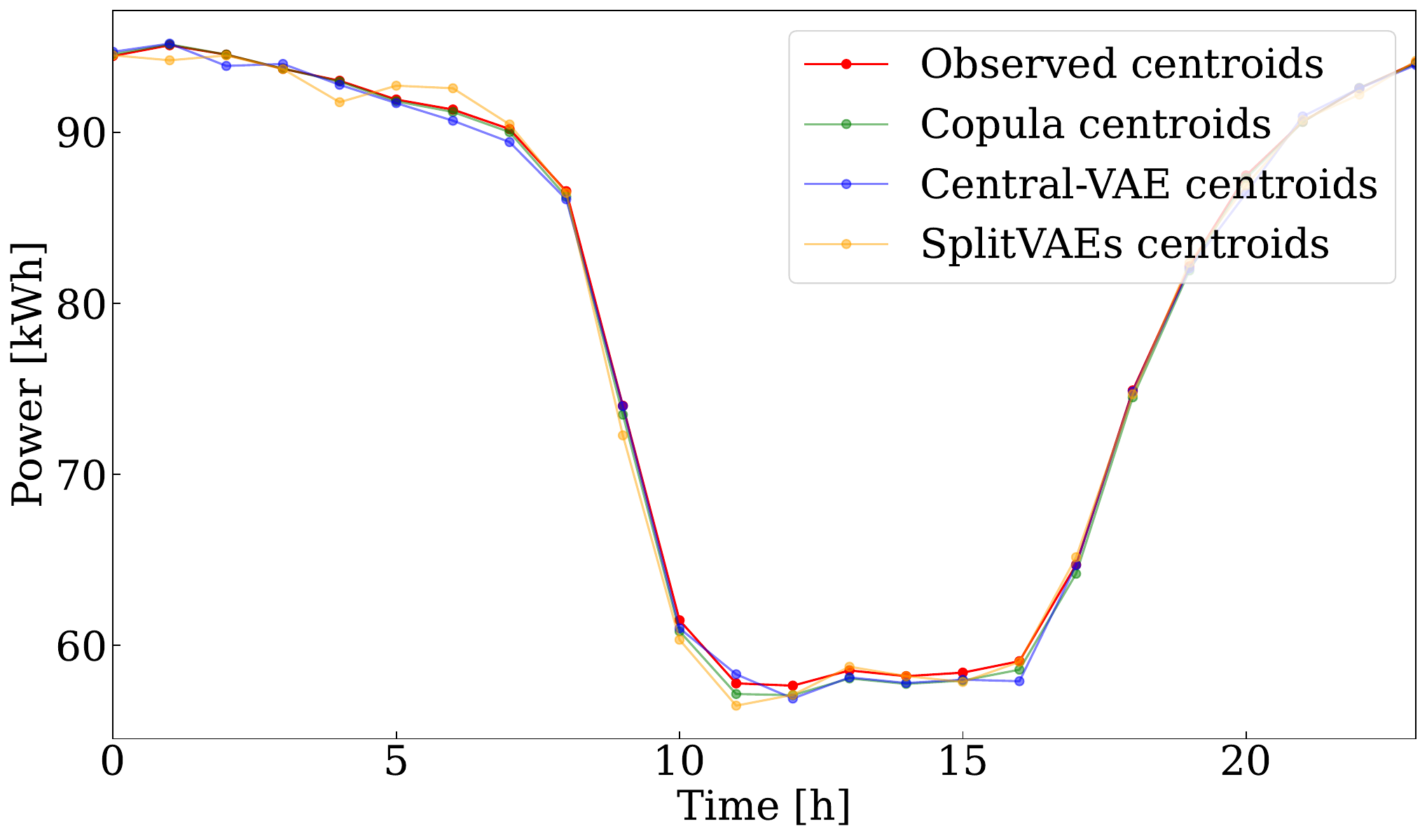}}
  \caption{Comparison of centroids between observed and generated data across four datasets.}\label{fig:tsa_centroid}
  \vspace{-3mm}
\end{figure*}
\begin{figure*}[htbp]
    \centering
        \subfigure[ACES Autocorrelation]{\includegraphics[width=0.32\textwidth,keepaspectratio]{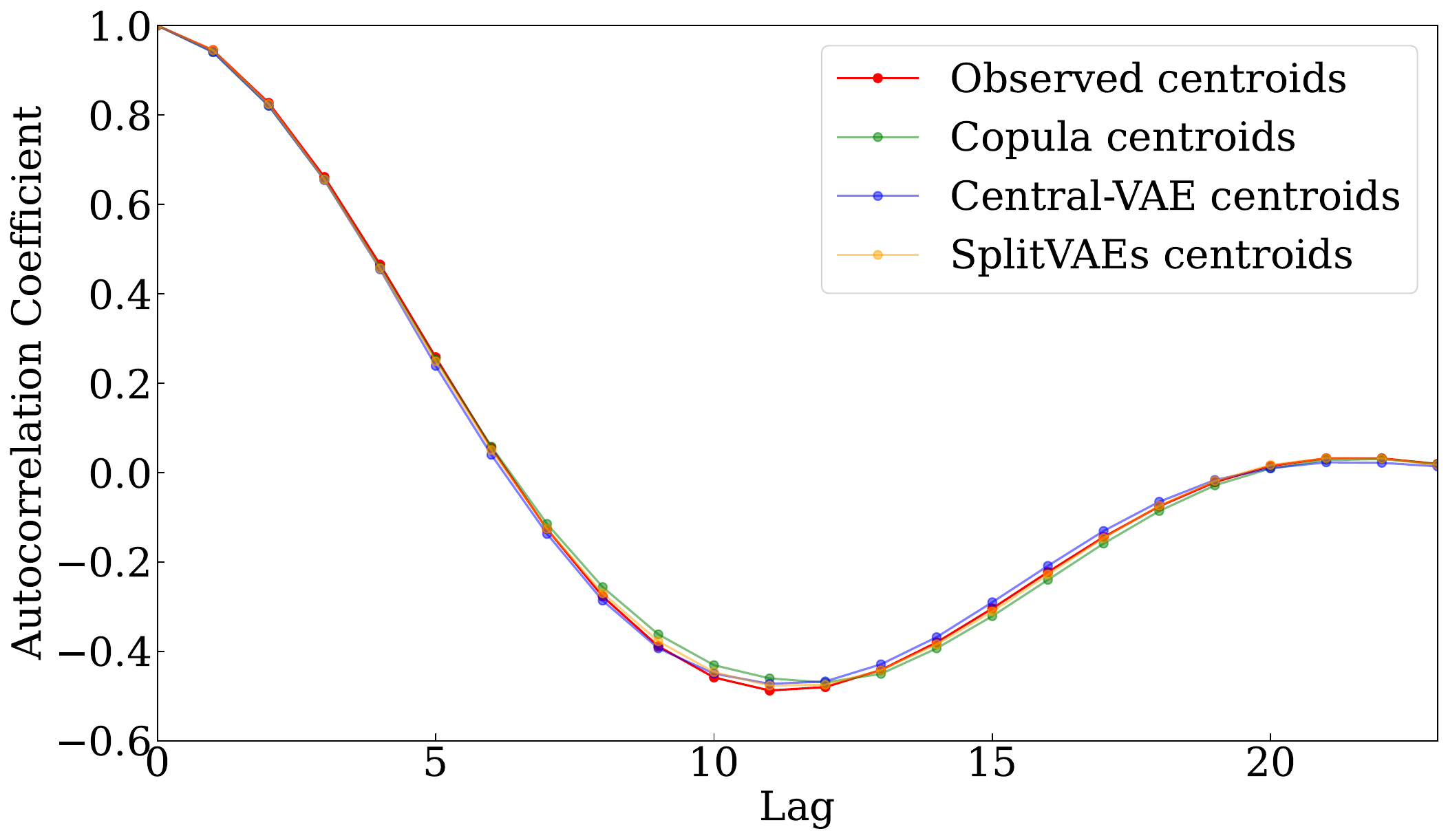}}
        \subfigure[Demand Autocorrelation]{\includegraphics[width=0.32\textwidth,keepaspectratio]{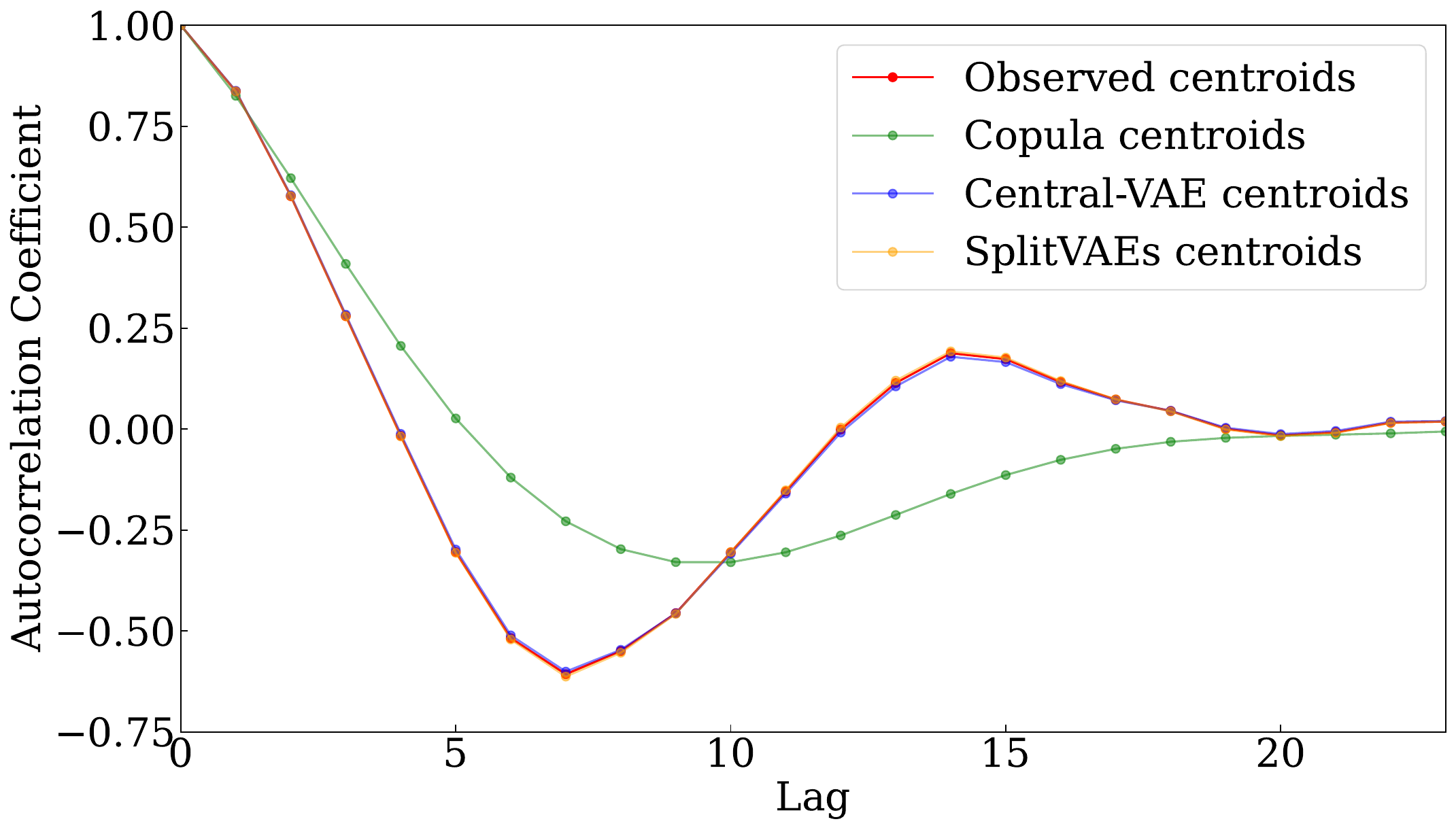}}
        \subfigure[Renewable Autocorrelation]{\includegraphics[width=0.32\textwidth,keepaspectratio]{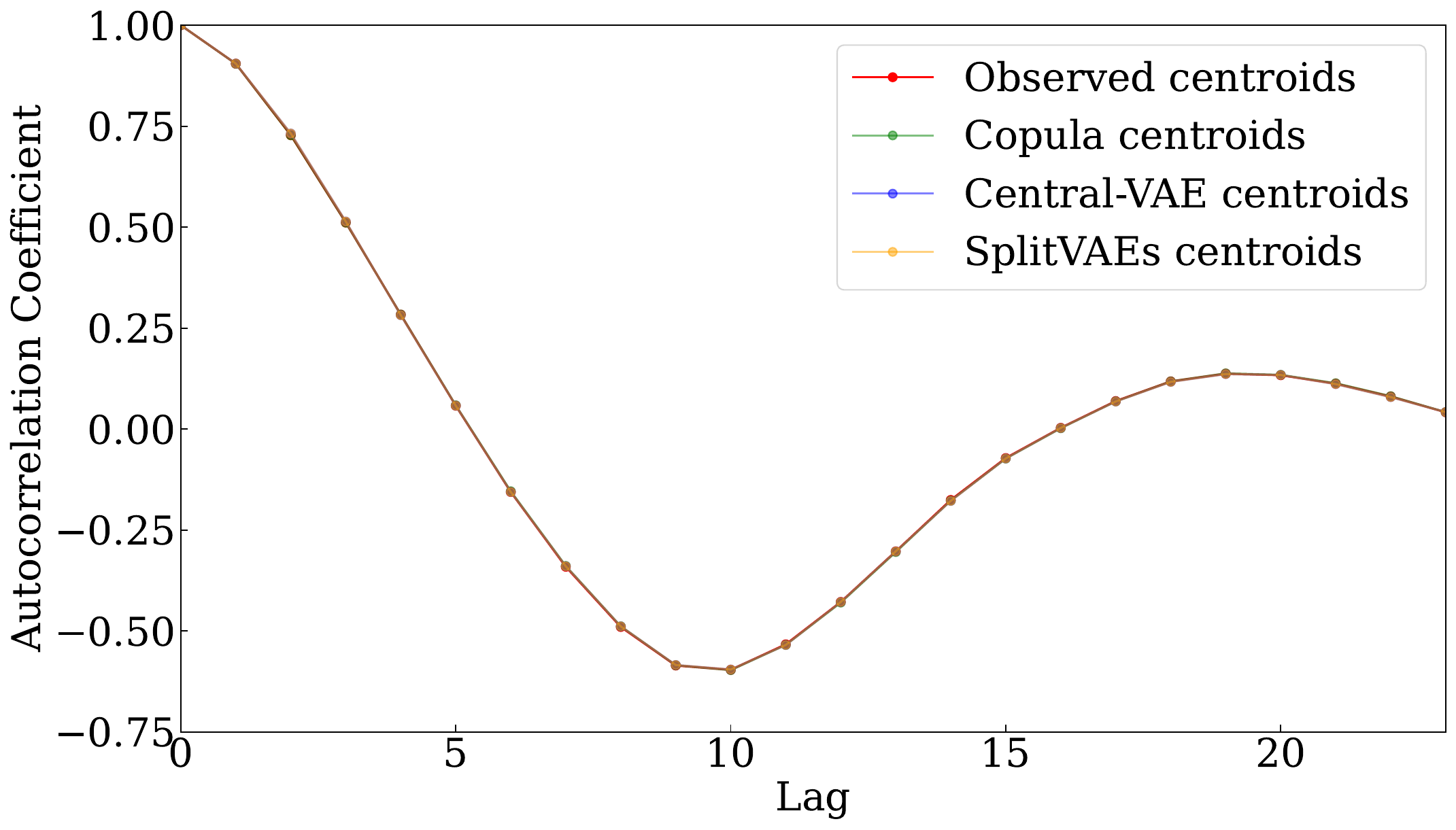}}
  \caption{Comparison of autocorrelations between observed and generated data across four datasets.}\label{fig:tsa_autocorrelation}
  \vspace{-3mm}
\end{figure*}
\begin{figure*}[]
    \centering
        \subfigure[Comparison of temporal information (each gray line represents a single-day scenario of emission volume generated by a particular method).]{\includegraphics[width=1.0\linewidth,height=0.25\linewidth]{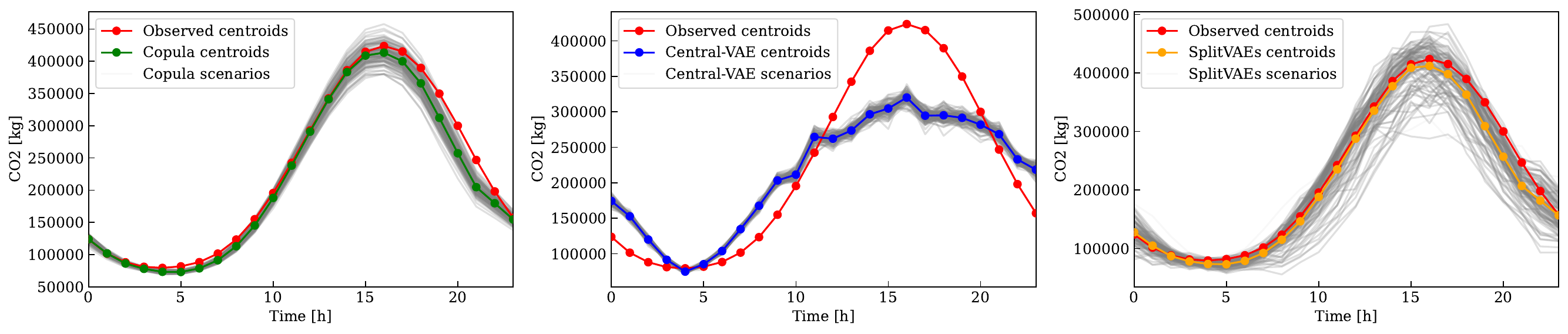}}
        \subfigure[Comparison of spatial information (each grid represents a $1\text{km}^2$ resolution of mean emission volume computed over 92 days).]{\includegraphics[width=0.95\textwidth,keepaspectratio]{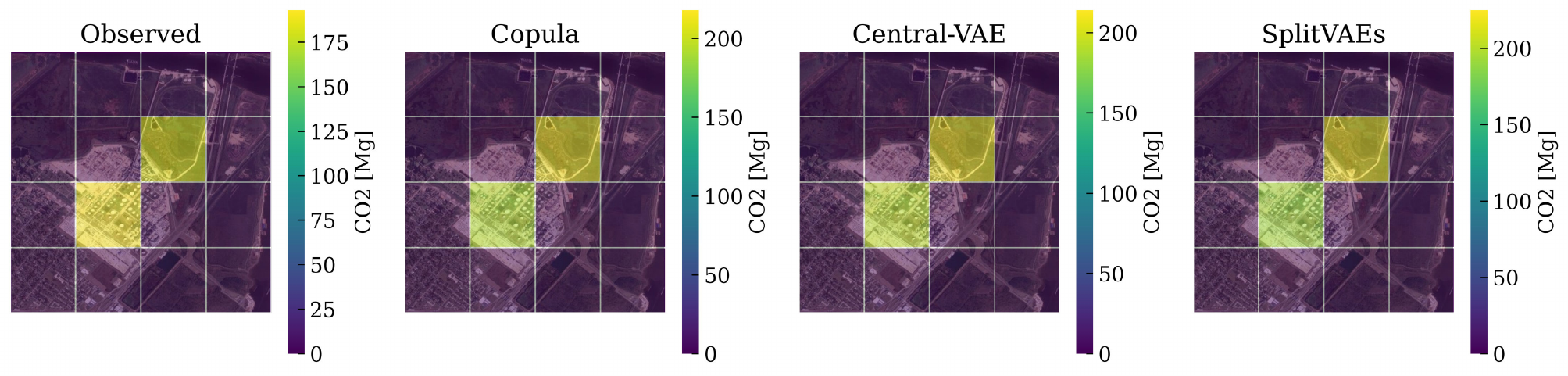}}
  \caption{A breakdown analysis of the spatial and temporal information extracted from generated scenarios at Port Arthur, Texas.}\label{fig:PortArthur}
  \vspace{-3mm}
\end{figure*}


\section{Experimental Results}\label{sec:results}
Our experimental results are primarily geared towards establishing the computational efficiency of the SplitVAEs method while delivering scenarios of the same statistical quality as the underlying dataset. In this section, we demonstrate that SplitVAEs deliver scenarios possessing similar statistical qualities to the observed data, while showcasing their ability of handling heterogeneous datasets in a scalable fashion.


\subsection{Hardware and Software Environment}
All experiments were conducted using the Pete High Performance Computing infrastructure from Oklahoma State University (OSU). All experiments pertaining to large-scale distributed processing were performed using the OpenMPI library \cite{gabriel2004open} coupled with Python bindings facilitated by \texttt{mpi4py} \cite{dalcin2021mpi4py}. We utilize the PyTorch library \cite{paszke2019pytorch} for constructing, training and evaluating the DNN models, and the inbuilt Ray Tune framework \cite{liaw2018tune} for hyperparameter tuning. Specifically, we first tune the hyperparameters of the Centralized-VAE, then adapt these hyperparameters for different numbers of edge locations and desired latent dimensions in SplitVAEs.

\subsection{Description of Key Datasets}
To demonstrate the broad applicability of our proposed framework, we primarily evaluate the decentralized scenario generation capability of our proposed framework using the USAID \cite{usaid}, ACES \cite{gately1943anthropogenic} and ACTIVSg datasets \cite{9174809}.  The USAID dataset comprises of time series data of health commodity supplies for different countries. The ACES dataset comprises hourly carbon dioxide ($\text{CO}_2$) emissions from the combustion of fossil fuels at a 1$\text{km}^2$ resolution in the continental United States from 2012 to
2017. In the ACES dataset, we extract time series data for emissions from 25 refineries in the US Gulf Coast during Summer 2017. Lastly, the ACTIVSg dataset provides the Demand and Renewable generation data for a power transmission network with an overall footprint for a large section of the state of Texas. We focus on hourly demand of 1125 individual network buses and renewable production from 87 renewable generators respectively which are referred to as Demand and Renewable datasets respectively.

\subsection{Applicability to a Diversity of Use Cases}\label{sec:results-diversity}
Table \ref{4cases} presents the mean and standard deviation of each evaluation metric for 100 iterations, with output dimension of edge level autoencoders set to 20. The experiments for each dataset comprises varying number of edge nodes ranging from 25 for ACES to 1125 distinct edge processes in the Demand dataset.
From Table \ref{4cases}, we can clearly observe that the performance of our proposed framework is nearly identical to those of centralized training schemes. Thus, we can expect SplitVAEs to generate high-fidelity scenarios for different application areas including but not limited to supply chains, carbon credit compliance planning and forecasting renewable energy production for power systems.

\begin{table}[htbp]
\caption{Means (with standard deviations measured in $10^{-3}$) of various evaluation scores across four different datasets.}
\begin{center}
\vspace{-3mm}
\begin{tabular}{p{1.0cm}p{0.5cm}cccc}
\hline
\multirow{3}{*}{Dataset}   & \multirow{3}{*}{\begin{tabular}[c]{@{}c@{}}Edge \\ Nodes\end{tabular}} & \multirow{3}{*}{Metric} & \multicolumn{1}{c}{\multirow{3}{*}{Copula}} & \multicolumn{1}{c}{\multirow{3}{*}{Central-VAE}} & \multicolumn{1}{c}{\multirow{3}{*}{SplitVAEs}} \\
                           &                                                                          &                         & \multicolumn{1}{c}{}                        & \multicolumn{1}{c}{}                             & \multicolumn{1}{c}{}                           \\
                           &                                                                          &                         & \multicolumn{1}{c}{}                        & \multicolumn{1}{c}{}                             & \multicolumn{1}{c}{}                           \\ \hline
                           \multirow{4}{*}{ACES}      & \multirow{4}{*}{\textbf{25}}                                             & FID                     & 2.095 (7)                                   & 2.255 (8)                                        & 2.397 (11)                                     \\
                           &                                                                          & ES                      & 0.122 (3)                                   & 0.124 (4)                                        & 0.136 (9)                                      \\
                           &                                                                          & RMSE                    & 0.201 (18)                                  & 0.218 (16)                                       & 0.304 (20)                                     \\
                           &                                                                          & CRPS                    & 0.209 (8)                                   & 0.2327 (14)                                      & 0.2649 (17)                                    \\ \hline
\multirow{4}{*}{USAID}     & \multirow{4}{*}{\textbf{85}}                                             & FID                     & 2.177 (7)                                   & 2.752 (7)                                        & 2.814 (6)                                      \\
                           &                                                                          & ES                      & 0.092 (1)                                   & 0.164 (1)                                        & 0.186 (1)                                      \\
                           &                                                                          & RMSE                    & 0.240 (1)                                   & 0.280 (1)                                        & 0.287 (2)                                      \\
                           &                                                                          & CRPS                    & 0.168 (1)                                   & 0.250 (2)                                        & 0.280 (10)                                     \\ \hline
\multirow{4}{*}{Renewable} & \multirow{4}{*}{\textbf{87}}                                             & FID                     & 2.105 (9)                                   & 2.167 (8)                                        & 2.169 (4)                                      \\
                           &                                                                          & ES                      & 0.307 (6)                                   & 0.309 (4)                                        & 0.310 (6)                                      \\
                           &                                                                          & RMSE                    & 0.309 (5)                                   & 0.334 (10)                                       & 0.351 (14)                                     \\
                           &                                                                          & CRPS                    & 0.320 (6)                                   & 0.322 (4)                                        & 0.329 (8)                                      \\ \hline

\multirow{4}{*}{Demand}    & \multirow{4}{*}{\textbf{1125}}                                           & FID                     & 1.540 (9)                                   & 1.545 (16)                                       & 1.947 (18)                                     \\
                           &                                                                          & ES                      & 0.129 (9)                                   & 0.143 (17)                                       & 0.148 (18)                                     \\
                           &                                                                          & RMSE                    & 0.328 (8)                                   & 0.419 (15)                                       & 0.462 (19)                                     \\
                           &                                                                          & CRPS                    & 0.180 (3)                                   & 0.182 (5)                                        & 0.184 (11)                                     \\ \hline

\end{tabular}\label{4cases}
\end{center}
\vspace{-3mm}
\end{table}
\begin{figure}[!htb]
    \centering
    \includegraphics[width=0.43\textwidth,keepaspectratio]{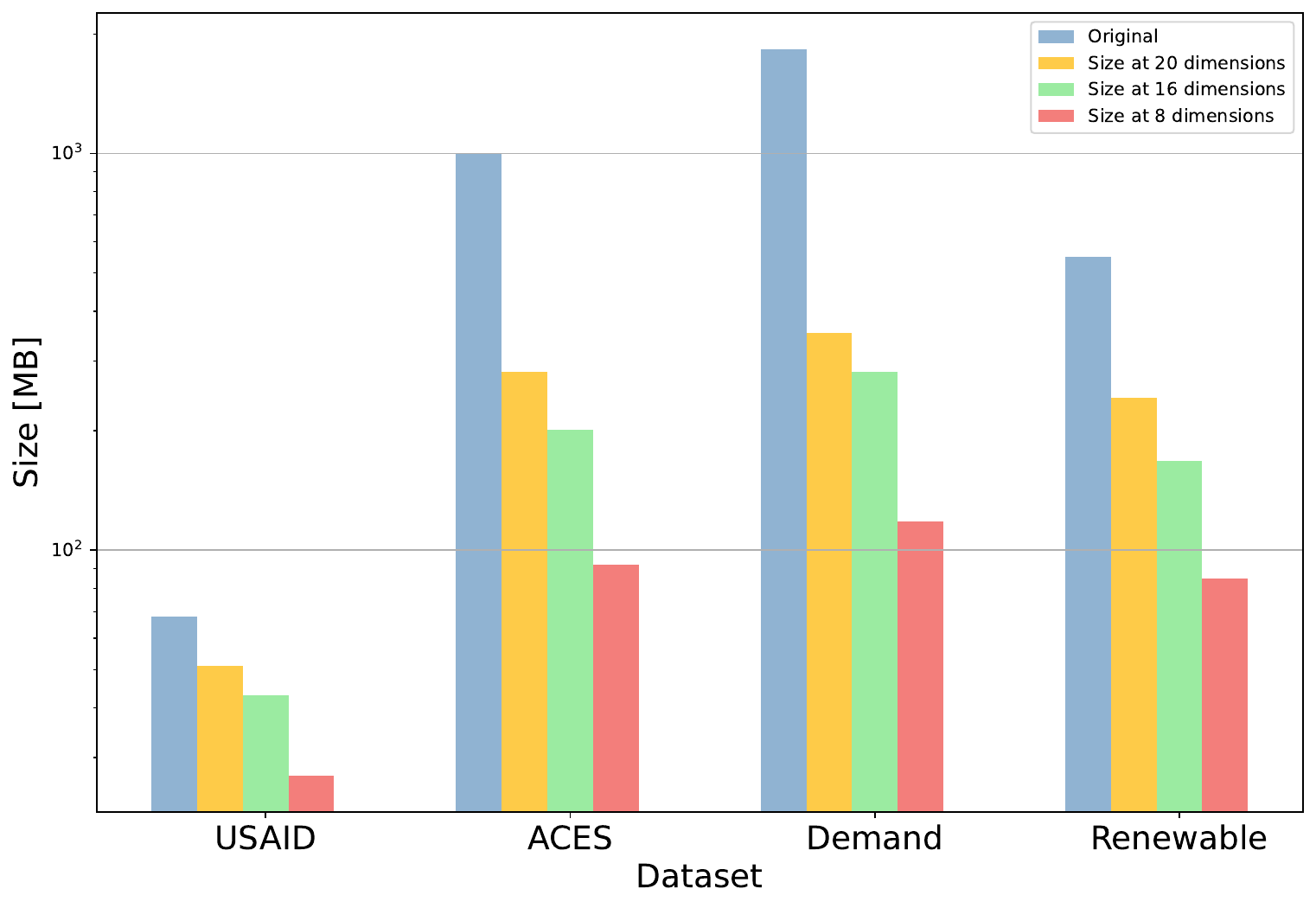}
    \caption{Analysis of reduction in transmitted data size, measured in log-scale, across different latent dimensions.}
    \label{fig:data_transfer}
    \vspace{-3mm}
\end{figure}

\begin{figure*}[!htb]
    \centering
        \subfigure[20 Regions]{\includegraphics[width=0.32\textwidth,keepaspectratio]{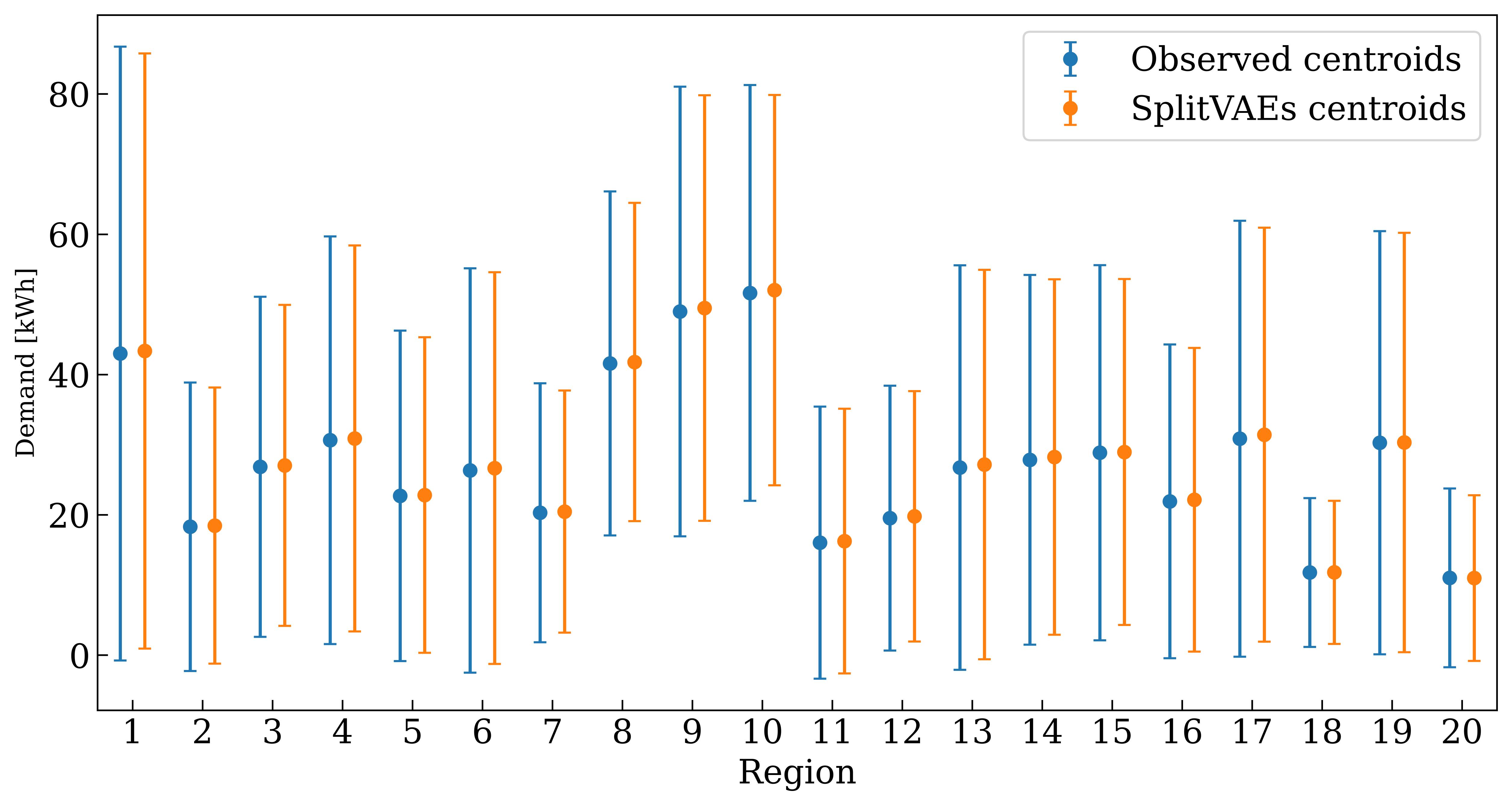}\label{fig:decomposition_20_regions}}
        \hfill
        \subfigure[60 Regions]{\includegraphics[width=0.32\textwidth,keepaspectratio]{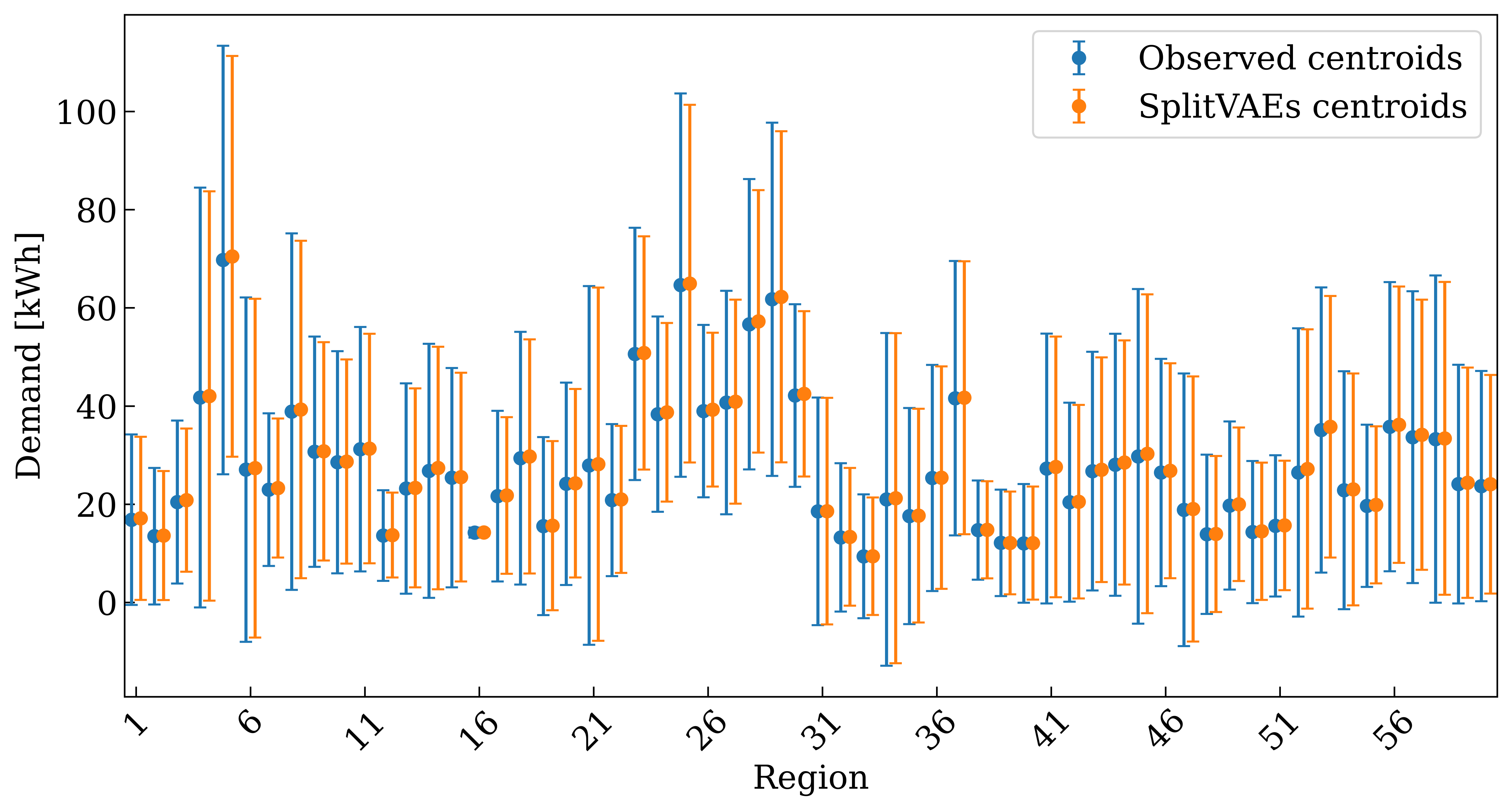}\label{fig:decomposition_60_regions}}
        \hfill
        \subfigure[120 Regions]{\includegraphics[width=0.32\textwidth,keepaspectratio]{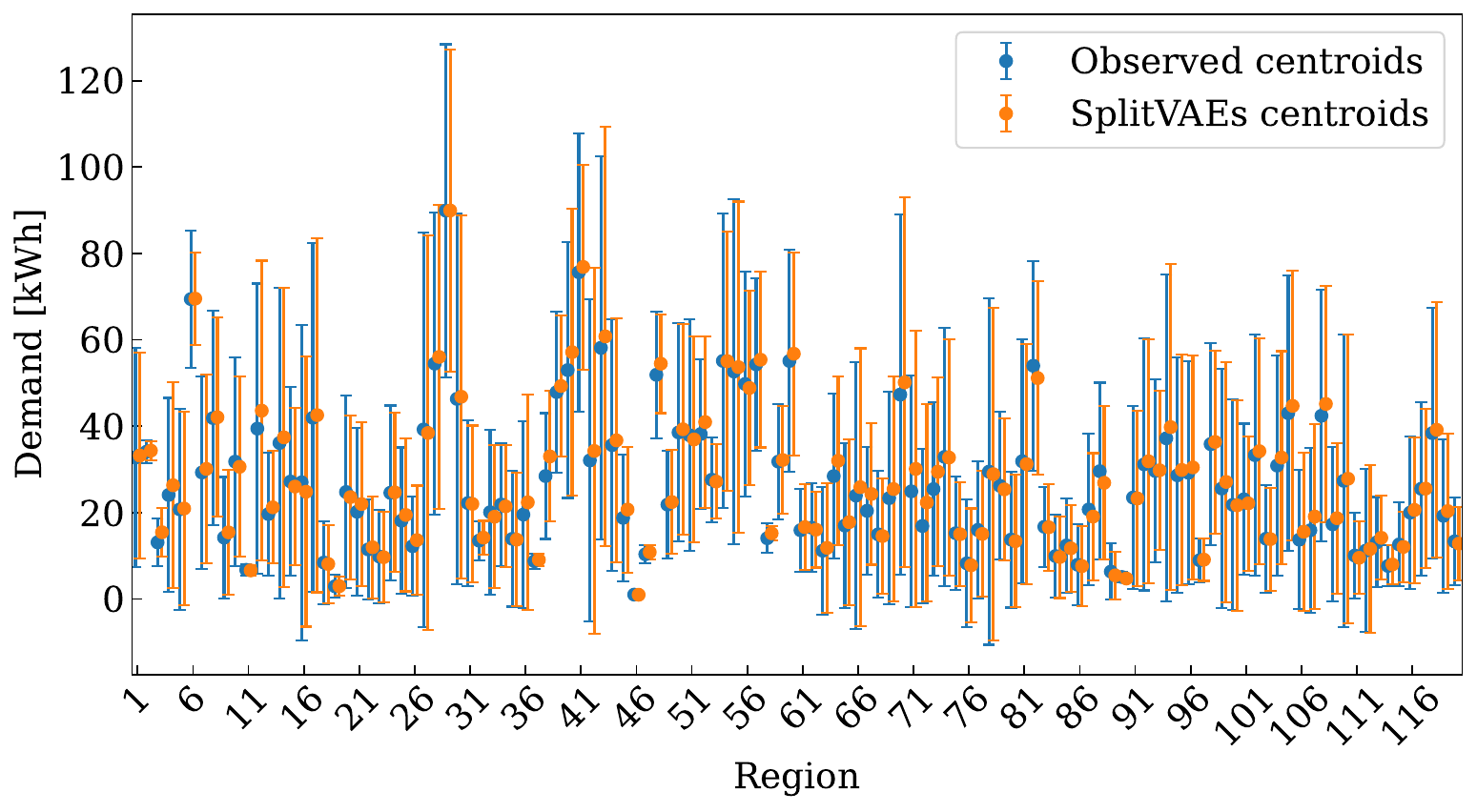}\label{fig:decomposition_120_regions}}
  \caption{An analysis of scenarios generated using three different region decompositions, compared with the observed data.}\label{fig:decomposition_spatiotemporal}
\end{figure*}

\begin{figure}[!htb]
    \centering
    \includegraphics[width=0.45\textwidth,keepaspectratio]{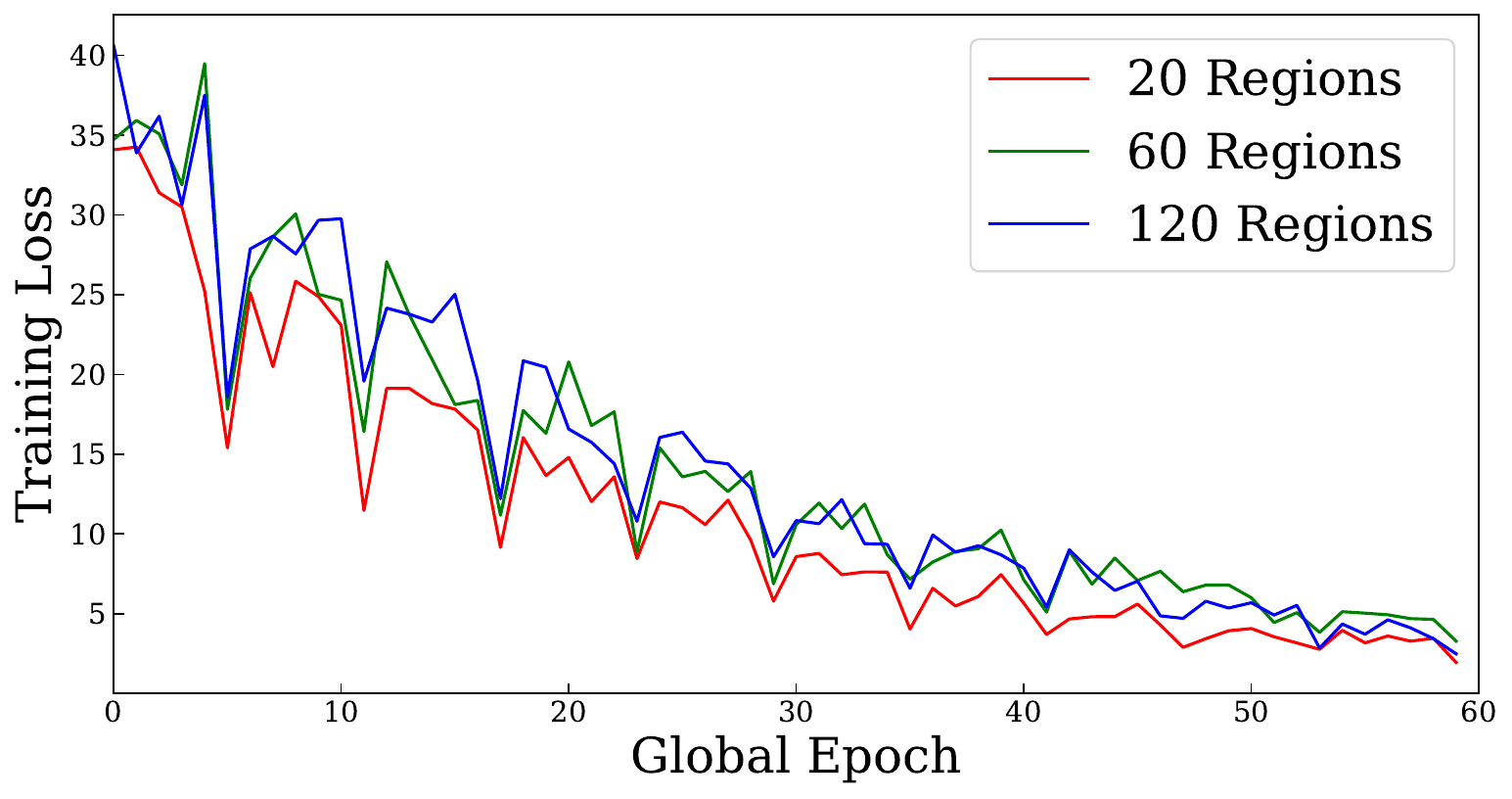}
    \caption{Global training losses across various settings of region decompositions in the ACTIVSg-Demand dataset.}
    \label{fig:decomposition_loss}
    \vspace{-3mm}
\end{figure}

Additionally, Figures \ref{fig:t-SNE}, \ref{fig:tsa_centroid}, and \ref{fig:tsa_autocorrelation} present a comparison of the observed data with scenarios generated by three different methods across four case studies. Figure \ref{fig:t-SNE} illustrates density curves, estimated using a Gaussian-based kernel method, of one-dimensional embeddings constructed from generated scenarios and observed data by the t-SNE method \cite{Hinton_2008}. Meanwhile, Figures \ref{fig:tsa_centroid} and \ref{fig:tsa_autocorrelation} provides a comparison of centroids - computed by averaging the quantity of interest  for all the nodes at each distinct time point - and their corresponding autocorrelation coefficients, respectively. 

From Figure \ref{fig:t-SNE}, we can observe that all methods capture the underlying distributions well, with 95\% of their principal components falling within the confidence range. Some minor misalignments, primarily result from feature compression in the latent dimension, do not significantly affect the overall alignment between the generated scenarios and the observed data. Further, Figures \ref{fig:tsa_centroid} and \ref{fig:tsa_autocorrelation}, show a close alignment between scenarios generated by SplitVAEs and state-of-the-art benchmark methods while capturing interdependencies present in original datasets both in terms of centralized and autocorrelations. 

To illustrate a practical application, we present seasonal carbon emission scenarios from a refinery at Port Arthur, Texas, in Figure \ref{fig:PortArthur}. As shown, the scenarios generated by the SplitVAEs closely align with the observed data and other benchmark methods. Specifically, within the perimeter of the industrial complex, the observed emission profile is approximately 180 to 190 Megagrams (Mg), which is accurately reflected in the scenarios generated by all three methods. Moreover, the emission patterns derived from the scenarios generated by the SplitVAEs follow the same trend as the observed data.  

\subsection{Reduction in Data Movement}

To evaluate the communication or transmission efficiency of SplitVAEs, we measure the total payload size handled by the system after a full cycle of our Algorithm \ref{alg:vae} comprising one forward pass of embeddings followed by backward pass of gradients and errors. Figure \ref{fig:data_transfer} compares the original dataset size with the transmission overhead for various edge dimension choices. The results show a significant reduction in data size as latent dimensions decrease. For instance, in the Demand dataset, the transmission size drops from 1830 MB for the original data to 475 MB at latent size 20, 223 MB at size 16, and 154 MB at size 8. Data transmission reduction factors at the smallest dimensions are $4.7$, $2.3$, $2.8$, and $2.7$ for the USAID, ACES, Demand, and Renewable datasets, respectively with respect to original data sizes. Combined with Figure \ref{fig:t-SNE}, Figure \ref{fig:tsa_autocorrelation}, and Table \ref{4cases}, these findings demonstrate that SplitVAEs deliver high-fidelity scenarios with up to nearly a 5 times reduction in data transmission costs.


\subsection{Handling Dimensional Heterogeneity}
In this case study, we evaluate the ability of SplitVAEs to handle heterogeneous data dimensions present across various stakeholders. Dimensional heterogeneity refers to differences in the size and structure of datasets across various edge devices in a decentralized network. For our experiments, we decompose the Demand dataset into regions with varying numbers of power system buses, where each bus represents a unique temporal demand curve. This results in multiple region decomposition cases, with each region acting as a distinct edge device handling a different dimensional dataset. Our decomposition cases rely on graph-partitioning schemes \cite{karypis1997metis} that attempt to decompose the power network graph by minimizing the total number of inter-regional transmission lines. As a result, we obtain three different decompositions pertaining to $n=20$ regions, $n=60$ regions, and $n=120$ regions wherein, each region represents a distinct SplitVAE edge node and the dimension of local data at each node is non-homogeneous. 


Figure \ref{fig:decomposition_spatiotemporal} provides a visual representation of the data, depicted as means (solid circle) and standard deviations (whiskers), generated by the SplitVAEs model across the three different regional decompositions. The plots demonstrate that the model can adapt to varying local dataset dimensions, effectively capturing the underlying data distributions in all cases. Therefore, the results suggest that the SplitVAEs framework is robust to dimensional heterogeneity, delivers high performance across different regional decompositions. On the other hand, Figure \ref{fig:decomposition_loss} illustrates the global training loss for the SplitVAEs model across different regional decompositions of the Demand dataset. The x-axis represents the number of global training epochs, while the y-axis shows the total training loss, which includes both reconstruction loss and KL divergence loss. Each line in the graph corresponds to a different regional decomposition setting. As observed, all trends show initial losses that decrease over subsequent epochs, indicating the model’s convergence.

\begin{figure}[!htbp]
    \centering
    \subfigure[Latent dimensions]{\includegraphics[width=0.45\textwidth,keepaspectratio]{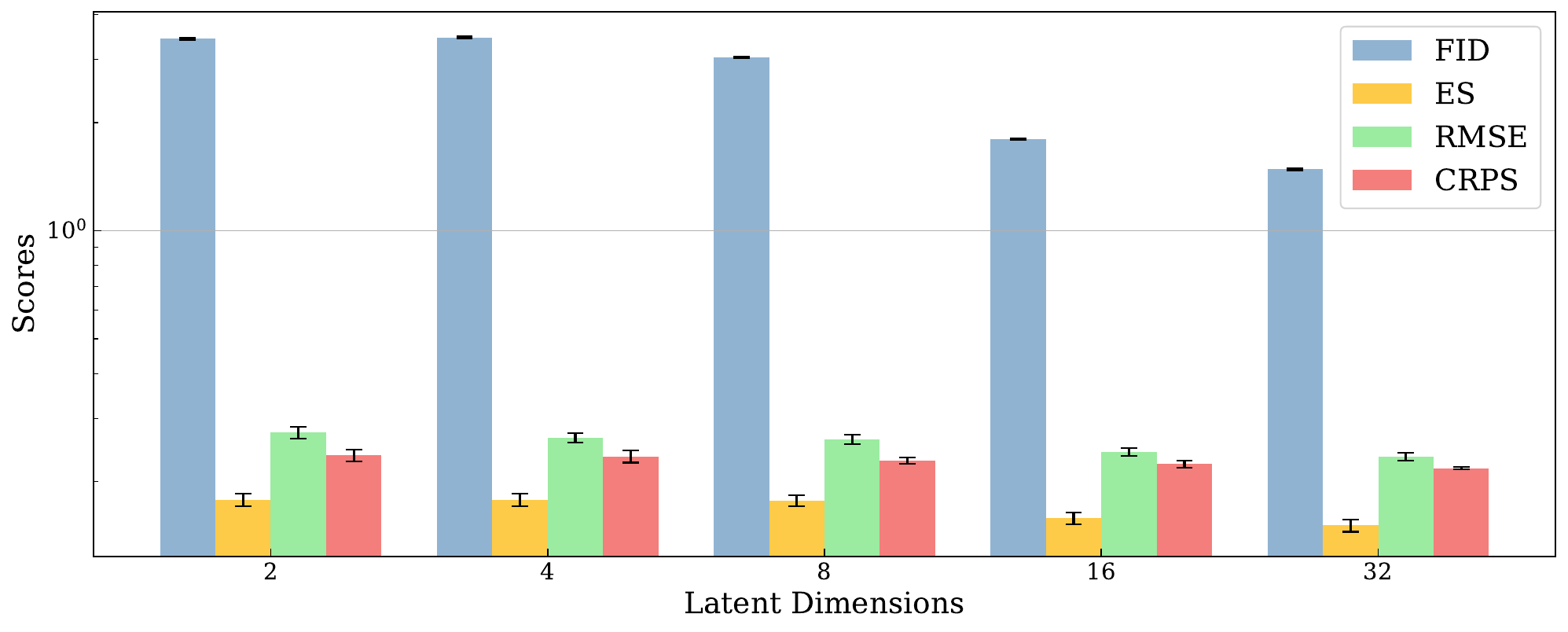}}\label{fig:scale_server}\\
    \subfigure[Edge-level output dimensions]{\includegraphics[width=0.45\textwidth,keepaspectratio]{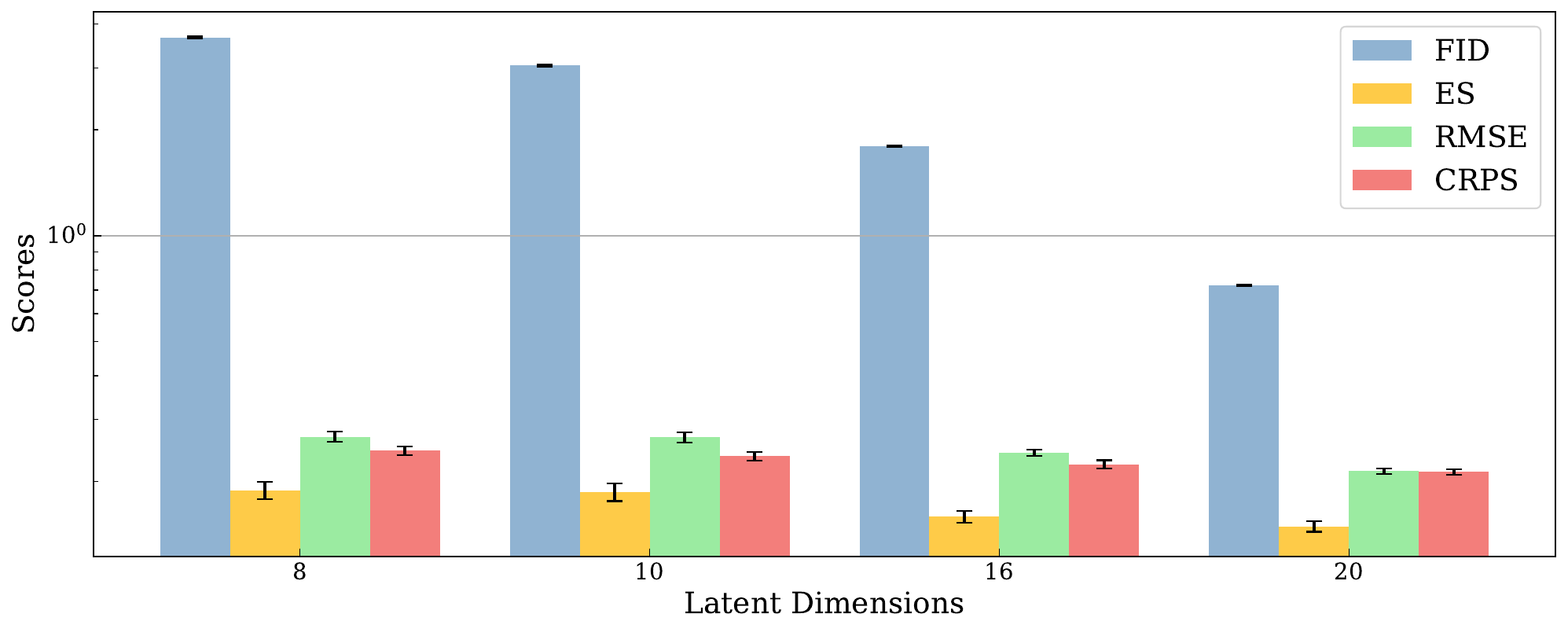}}\label{fig:scale_edge}
    \caption{Analysis of the SplitVAEs architecture's robustness on ACTIVSg Demand, presented in log-scale.}\label{fig:scalability}
    \vspace{-1mm}
\end{figure}

\subsection{Architectural Robustness}
In this case study, we aim to evaluate the robustness of SplitVAEs with varying dimensions of edge-level AE outputs and server-level VAE latent embeddings across all four evaluation metrics. For each experiment, we consider 100 scenario generation instances employed to provide a statistically sound set of scores for evaluation. Our results are presented in Figure \ref{fig:scalability} with respect to the Demand dataset, where means and standard deviations are depicted as bars and whiskers, respectively. In Figure \ref{fig:scalability}a, we observe that an increase in the VAE latent embedding size, from 2 to 32, greatly improves the performance of the SplitVAEs, as evidenced by lower scores across all metrics. Notably, the FID scores exhibit a significant improvement, indicating that models with larger latent dimensions are generating outcomes that closely resemble trends in the original dataset. Meanwhile, Figure \ref{fig:scalability}b demonstrates that increasing AE output sizes in general leads to a higher quality of scenarios as expected. Overall, the trends presented in Figure \ref{fig:scalability} conclusively show that the performance of SplitVAEs remains robust to varying degree of latent dimensions across both server and edge models.

\section{Conclusion}\label{sec:conclusion}
In the paper we present SplitVAEs, a novel decentralized method to generate spatiotemporally interdependent scenarios from siloed data specifically geared toward solving SO problems. Using backpropagation, we effectively decompose the training across edge-level autoencoders and a server-level variational autoencoder and enable the bi-directional flow of insights among the edge and server models. We demonstrate the applicability of SplitVAEs specifically in the context of multi-stakeholder infrastructure systems by leveraging real-world datasets from various areas including supply chains, energy, and environmental science. We show that SplitVAEs generate scenarios that closely align with the real underlying data distributions by benchmarking with the state-of-the-art centralized methods and numerous statistical metrics. Using large scale, distributed memory driven experiments, our results indicate that SplitVAEs can significantly reduce data transmission, scale efficiently with growing number of edge locations and help break down data silos. Overall, our analysis demonstrates that SplitVAEs are a compelling alternative that enables scenario generation for SO problems in multi-stakeholder-led infrastructure systems without the need to transfer underlying datasets.


\end{document}